%% file: paper.tex
\documentclass[]{fairmeta}

\usepackage{hyperref}       %
\usepackage{url}            %
\usepackage{booktabs}       %
\usepackage{amsfonts}       %
\usepackage{nicefrac}       %
\usepackage{microtype}      %
\usepackage{xcolor}         %
\usepackage{graphicx}
\usepackage{todonotes}
\usepackage{multirow}
\usepackage{tcolorbox}
\usepackage{float}
\usepackage{hyperref}       %
\usepackage{amsmath}
\usepackage{cleveref}
\usepackage{enumitem}
\usepackage{algorithm}
\usepackage{algpseudocode}

\AddToHook{cmd/appendix/before}{%
    \crefalias{section}{appendix}%
    \crefalias{subsection}{appendix}
}

\floatstyle{plaintop} 
\newfloat{promptbox}{t}{lop}
\floatname{promptbox}{Box}
\crefname{promptbox}{Box}{Boxes}
\Crefname{promptbox}{Box}{Boxes}
\crefname{tcb@cnt@tcolorbox}{Box}{Boxes}
\Crefname{tcb@cnt@tcolorbox}{Box}{Boxes}

\newcommand{\methodname}{ParaEval}

\title{Are We Evaluating Knowledge or Phrasing? Mitigating MCQA Sensitivity with \methodname{}}

\author[1,2]{João Maria Janeiro}
\author[1]{Mathurin Videau}
\author[1]{Andrea Caciolai}
\author[2]{Benjamin Piwowarski}
\author[2,3]{Patrick Gallinari}
\author[1]{Loic Barrault}

\affiliation[1]{FAIR at Meta}
\affiliation[2]{Sorbonne Université, CNRS, ISIR, F-75005 Paris, France}
\affiliation[3]{Criteo AI Lab, Paris, France}

\abstract{\input{abstract}}

\date{\today}
\correspondence{João Maria Janeiro at \email{joaojaneiro@meta.com}}

\begin{document}

\maketitle

\input{intro}
\input{related_work}
\input{preliminaries}
\input{analysis}
\input{method}
\input{results}
\input{conclusion}

\clearpage
\newpage
\bibliographystyle{assets/plainnat}
\bibliography{paper}

\clearpage
\newpage
\beginappendix

\input{appendix_ablations}

\clearpage\newpage

\end{document}

%% file: abstract.tex
\begin{abstract}

Multiple-choice (MCQA) benchmarks are the standard for evaluating pretrained large language models, but their reliance on log-likelihood scoring makes them unreliable. 
Specifically, standard scores are highly sensitive to the exact phrasing (surface form) of the answers, conflating a model's familiarity with a specific phrase with its actual capability.
We demonstrate this flaw using a controlled testbed of 1B-8B models trained on the same knowledge.
Despite having identical knowledge, standard metrics falsely report a performance gap of over 2 points. 
To solve this, we propose \textit{\methodname{}}, an evaluation framework that queries models using multiple paraphrases per answer option. 
By scoring each model based on its most favorable phrasing, \textit{\methodname{}} successfully reduces the false performance gap to below 1 point. 
We confirm that these evaluation artifacts, and the improvements from \textit{\methodname{}}, persist in frontier 70B and 120B open-source models. 
Ultimately, \textit{\methodname{}} provides a robust and efficient way to evaluate true underlying capability rather than surface-form familiarity.
\end{abstract}

%% file: intro.tex
\section{Introduction}
Benchmarks are how we measure knowledge in large language models.
They are taken as the source of truth for measuring the quality, knowledge and capabilities of current LLMs.
In practice, evaluation has converged on two paradigms: generative tasks where models write free-form answers, and multiple-choice question answering (MCQA) where models select from a predetermined set of options.

Generative evaluation is natural but expensive.
It requires autoregressive decoding, which is slow at scale and especially noisy for smaller models, and exact match against a single reference is brittle by design. In the LLM era, the field has largely converged on LLM-as-a-judge to overcome the later, which further exacerbate costs, show poor test-retest reliability, and exhibit systematic biases toward verbosity and style \citep{schroeder2025trust,soumik2026judging,wang2024large}.

A simpler, cheaper alternative is MCQA.
Scoring is a single forward pass per option, making it cheap, deterministic, and comparable across model sizes.
This efficiency made MCQA one of the main drivers of model development and performance measurement, but it is also famously brittle.
Because LLMs tackle these tasks as few-shot learners \citep{brown2020language}, their accuracy is notoriously sensitive to prompt design. Performance can swing wildly based on the order of demonstrations \citep{lu2022fantastically} or superficial formatting choices, which have been shown to alter scores by up to 76 points \citep{sclar2023quantifying}.
Brittlebench \citep{brittlebench} systematizes this instability, showing that semantics-preserving prompt perturbations degrade frontier models by up to 12\% and flip rankings in most comparisons.
Existing work has treated this as a problem of \textit{how to ask}.

\begin{figure*}[t]
\centering
\includegraphics[width=0.98\textwidth]{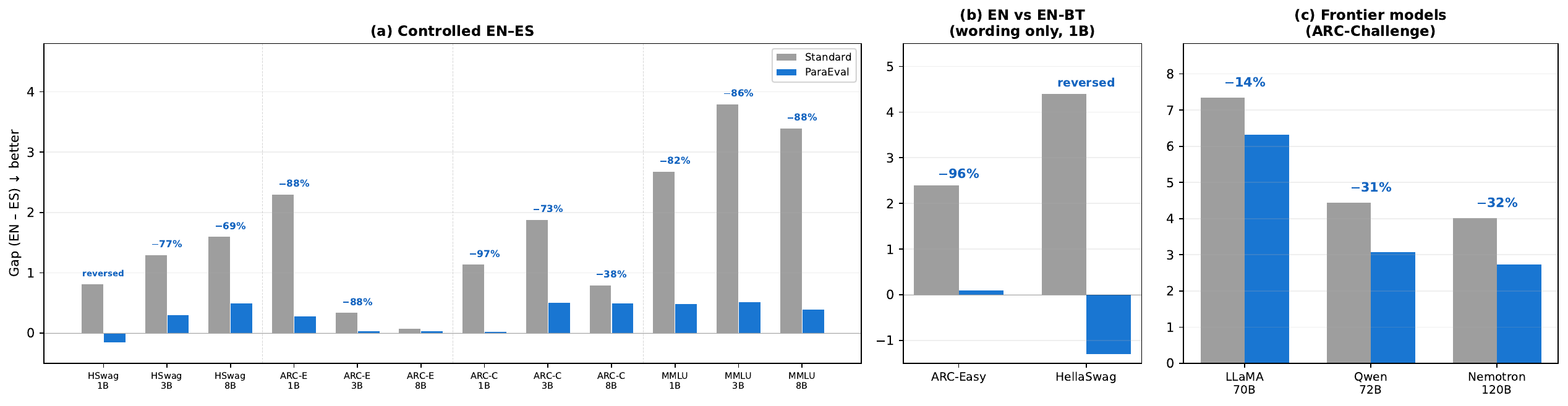}
\caption{\textbf{Overview of phrasing artifacts.} Standard MCQA scoring (gray) reports gaps between models with identical knowledge, while ParaEval (blue), which scores each answer option under multiple paraphrases, collapses those gaps. (a) Controlled English–Spanish models at 1B/3B/8B. (b) English vs. backtranslated-English training data at 1B isolates wording from language. (c) Large frontier models on ARC-Challenge show the same pattern. Full numbers are in \Cref{sec:results}.}
\label{fig:overview}
\end{figure*}

We argue the blind spot is \textit{what is scored}.
In MCQA most of the knowledge lives in the answer tokens themselves, the exact strings over which likelihood is computed.
A model trained with more Brazilian Portuguese might prefer \textit{onibus} to \textit{autocarro}, and the benchmark rewards whichever wording the author chose, conflating familiarity with knowledge.

To study phrasing sensitivity and model knowledge without confounding factors, we train monolingual English and Spanish decoder-only models (1B-8B) on the same exact underlying corpus.
We additionally train an English model on the same data paraphrased.
The only differences between training sets are language and wording, knowledge is held constant. If existing benchmarks were correct, these models should perform at the same level. They do not. \Cref{fig:overview} summarizes the gaps between models. We explore answer phrasing sensitivity on these models and expose gaps of more than 6 points by just rephrasing the correct answers. This brings into focus the reliability of the benchmarks. Are they really evaluating the knowledge of these models?

We therefore propose changing the current evaluation strategy to query through many surface forms of the answers.
ParaEval paraphrases all options, both correct and distractors, and scores each option by its highest-probability paraphrase.
For generation, we also augment the reference set with the same strategy.
With these protocols, the gap for models trained on identical knowledge is reduced substantially.
\Cref{fig:overview} shows the reductions under ParaEval, where all benchmarks are below 1\% absolute difference across all scales under identical knowledge, with up to 96\% reductions.
The pattern holds for strong open-source models at 70B and 120B scales, confirming that the artifact is not a small-model pathology.
This demonstrates that the underlying issue is not a lack of knowledge within the models, but more that evaluations query knowledge through a single arbitrary phrasing that may not match the model's distribution, hereby misjudging its true capabilities.

Our contributions are:
\begin{enumerate}
    \item We provide a controlled training framework of distinct, matched monolingual models. By comparing monolingual English, Spanish and backtranslated-English models, we isolate evaluation artifacts, revealing artificial performance gaps up to 4 despite models being trained on identical underlying knowledge.
    \item We demonstrate that standard MCQA scoring is sensitive to surface form: Paraphrasing only the correct answer raises accuracy by 5 to 8 points, and no common normalization removes this effect. Both our matched models at smaller scales (1B, 3B and 8B) are affected, but also frontier models at 70B and 120B scales (with similar performance changes under alternative phrasing).
    \item We propose ParaEval, a paraphrase-augmented MCQA method that scores each option by its best surface form, bringing models with identical knowledge to similar performance levels and reducing gaps by 14 to 96 percent across benchmarks at all scales.
    \item We show that multiple phrasings for generative evaluations using our method also help with gap reduction, eliminating the gap on Natural Questions and reducing it on TriviaQA.
\end{enumerate}

%% file: related_work.tex
\section{Related Work}

\textbf{Generative Evaluation.}
Open-ended generation is hard to evaluate given the degree of freedom that natural language possesses. The machine translation community first faced this problem, relying on lexical overlap against human references with metrics such as BLEU \citep{papineni2002bleu} and ROUGE \citep{lin2004rouge}. Given that semantically equivalent answers can have zero lexical overlap, they were later replaced by learned metrics like BLEURT \citep{sellam2020bleurt}. The field has now converged on LLM-as-a-judge \citep{zheng2023judging}, which is accurate but costly and biased toward style \citep{wang2024large, soumik2026judging}. Recent work
has listed further biases: judges favor responses in
specific positions regardless of quality \citep{wang2023not_fair}, 
exhibit self-enhancement bias toward their own outputs, and vary
dramatically in reliability across domains
\citep{tan2024judgebench}. 
MCQA sidesteps the expense and subjectivity of
generative evaluation entirely.

\textbf{MCQA Brittleness.}
MCQA remains standard for evaluations during training or under
budget constraints because it is cheap and deterministic, but it is
brittle to demonstration order \citep{lu2022fantastically},
calibration \citep{zhao2021calibrate}, and formatting
\citep{sclar2023quantifying}. Brittlebench \citep{brittlebench}
shows semantics-preserving prompt changes degrade accuracy up to
12\% and flip rankings. \citet{alzahrani2024benchmarks} demonstrate
that such perturbations, including option reordering and rephrasing, make
leaderboard rankings unstable. Models also exploit choice-only
shortcuts, relying on patterns such as synthetic distractors or detailed and more formal correct answers \citep{balepur2025which, chandak2025answer}.
Beyond ordering, LLMs exhibit \emph{selection bias}: a preference
for specific option symbols (e.g.\ ``A'') driven by token-level
priors \citep{zheng2024not_robust}. Scoring methodology itself
introduces variance: \citet{molfese2025right} show that  
first-token probability, full-text matching, and log-likelihood
scoring can disagree by over ten accuracy points on the same
benchmark. Reported multilingual gaps often reflect evaluation
artifacts rather than capability, as shown by large-scale audits
\citep{ahuja2023mega, ahuja2024megaverse} and by corrected
translations on MGSM \citep{peter2025mind}.

\textbf{Paraphrasing in LLM Evaluation.}
In this paper, we frame the aforementioned brittleness to surface-form, as concurrent work Brittlebench \cite{brittlebench} showcased. Similarly, \citet{holtzman2021surface} showed that probability mass splits across synonymous surface forms, depressing the score of the benchmark's chosen string. Reweighting via PMI or calibration does not reliably fix this and can harm instruction-tuned models \citep{wiegreffe2023increasing}.
As with generative evaluation, machine translation addressed this problem with human-paraphrased multi-references \citep{freitag-etal-2020-bleu}. Recent work paraphrases references for learned metrics \citep{tang2023not}, or paraphrases questions to test consistency \citep{jiang2020know, elazar2021measuring, choi2024roparq} and math problems to improve reasoning \citep{zhou2024paraphrase}. 
We extend this line of research by paraphrasing \emph{answer
options} in MCQA and scoring each option by its highest-probability
paraphrase.

%% file: preliminaries.tex
\section{A controlled testbed for evaluation}
\label{sec:analysis}

Isolating evaluation artifacts requires a testbed free of training confounds. We establish a controlled setup where models train on equivalent knowledge, despite exposure to different surface forms, either across languages or via alternative English phrasing. Consequently, if these matched models score differently under standard metrics, the resulting gap strictly isolates a measurement artifact.

\subsection{Preliminaries: Standard Evaluation Metrics}
\label{sec:preliminaries}

We study two dominant LLM evaluation paradigms, both of which are fundamentally tied to the specific surface forms of ground-truth answers.

\paragraph{Multiple-choice question answering (MCQA).}
Following \citet{brown2020language}, given a question $q$ and a set of candidate answers $\{a_1, \ldots, a_m\}$, the standard MCQA evaluation selects the answer with the highest log-likelihood:
\begin{equation}
\hat{y} = \mathop{\mathrm{argmax}}_{i} \, f(\log P(a_i \mid q))
\label{eq:mcqa_scoring}
\end{equation}
where $f$ is an optional normalization function, typically dividing by character count (\texttt{acc\_norm}) or token count. Because $\log P(a_i \mid q)$ is computed over the specific tokens in $a_i$, two semantically equivalent phrasings of the same answer can yield different scores. Words that appear less often in the model's training distribution will induce higher perplexity.

\paragraph{Generative exact match.}
Open-ended generation benchmarks prompt the model to generate a response and score it by exact match against a reference set $R = \{r_1, \ldots, r_n\}$. A response is correct if it matches any $r_j \in R$. The quality of this evaluation is strictly bounded by the completeness of $R$.

\subsection{A Controlled Testbed for Cross-Lingual Evaluation}
\label{sec:controlled_testbed}

Cross-lingual comparisons are typically skewed by disparities in training data and tokenization. To isolate evaluation deficiencies, we establish a strictly controlled baseline: English and Spanish models trained from scratch to possess equivalent knowledge acquired during training.

\textbf{Matched Data.}
We train decoder-only transformer language models (LLaMA architecture) of 1B, 3B, and 8B parameter. To equalize knowledge content, both the English and Spanish models are trained on the exact same underlying corpus: the original DCLM-EDU \citep{dclm_edu} for English and an LLM-translated version (using GPT-OSS \citep{gpt-oss}) of the exact same 120B tokens for Spanish.
Details and quality assessments are given in \Cref{section:appendix/translation_of_training_data}.

\textbf{Matched Tokenizers.}
The default LLaMA tokenizer produces higher fertility and lower compression for non-English text, meaning non-English models effectively see much less data at a fixed token budget. We eliminate this confounding factor by training language-specific tokenizers (128k vocabulary each) that achieve matched compression ratios ($\sim$5.0 chars/tok for English, $\sim$5.1 for Spanish). Moreover, for each data configuration all models are trained on the same hyperparameters (e.g., LR, batch size). More details can be found in \Cref{app:baselines}.

\textbf{Hyperparameters} We use consistent hyperparameters for each model scale across all languages, following DCLM recommendations \cite{li2024datacomp}. Full details are provided in appendix \cref{tab:hyperparameters}.

\textbf{Benchmarks} We translate all benchmarks into the target languages using Claude 4.6 Opus. Evaluations are conducted via the \texttt{lm-harness} framework \footnote{\url{https://github.com/EleutherAI/lm-evaluation-harness/}}, with further details available in \cref{section:appendix/training_setup}.

\subsection{The Wording Control: English vs. Backtranslated English}
\label{sec:wording_control}

Although controlling for tokenization and data reduces confounding factors, comparing English to Spanish still crosses a linguistic boundary. To definitively isolate \emph{phrasing} from \emph{language}, we introduce a symmetric set of secondary controls entirely within English. 
To explore the effect in English only, we train an additional 1B English model from scratch on 60B backtranslated tokens (EN$\to$ES$\to$EN) of the same DCLM-EDU corpus. These two English models (Standard and Backtranslated) possess identical document-level knowledge but differ in the stylistic distribution of their training data. By evaluating both models on standard English benchmarks, any performance gap perfectly isolates the effect of phrasing familiarity without introducing cross-lingual variables.

More details of training and evaluation are provided in \Cref{section:appendix/training_setup}, and the confounding controls are explored in detail in \Cref{app:baselines}.

%% file: analysis.tex
\section{Diagnosing Surface-Form Sensitivity}
\label{sec:diagnosing}

Despite controlling for all training confounding factors, (detailed in \Cref{app:baselines}), a residual gap of 4--6\% remains on most benchmarks across all scales. We therefore ask the question: \textbf{is this remaining gap a genuine capability difference, or is it an evaluation artifact?}

\subsection{MCQA as generation task and normalization effect}
To get a better assessment of whether MCQA results are affected by its evaluation protocol, we cast the ARC-Easy \citep{clark2018think} evaluation as a generative task.
In this generative version, models generate the answer given a 5-shot prompt in the native language that includes both questions and answers. 
Results are presented in \Cref{tab:scoring_methods}, where it shows English and Spanish models perform almost equally with the gap greatly reduced. This suggests that the models are at a level of similar knowledge, as we hypothesize, since the models were trained on identical knowledge.

We tested whether alternative scoring strategies on ARC-Easy could mitigate this evaluation bias (\Cref{tab:scoring_methods}). 
Raw log-likelihood (raw LL) yields a +5.3\% gap in favor of English. 
Normalizing by characters widens this to +5.9\%, and normalizing by tokens inflates the gap to +8.0\%. 
The PMI score, defined as $PMI(a_i|q) = logP(a_i|q) - logP(a_i)$, which better accounts for normalization and standalone fluency of the answer, exhibits a slightly reduced gap (+4.4\%).
Because the underlying issue is the failure to match the model's distributional expectations, no common normalization technique eliminates the gap.

\begin{table}[t]
\centering
\scriptsize{%
\caption{ARC-Easy different scoring methods (1B scale). English-only (EN) vs.\ Spanish-only (ES).}
\label{tab:scoring_methods}
\begin{tabular}{lccc}
\toprule
Method & EN & ES & $\Delta$ \\
\midrule
acc (raw LL)          & 71.6 & 66.3 & +5.3 \\
acc\_norm (chars)     & 66.6 & 60.7 & +5.9 \\
acc\_norm (tokens)    & 65.7 & 57.7 & +8.0 \\
PMI                   & 61.2 & 56.9 & +4.4 \\
\midrule
Generative (5-shot EM) & 33.1 & 33.2 & --0.1 \\
\bottomrule

\end{tabular}
}
\end{table}

\subsection{Prompt: language and few-shot}
Standard benchmarks often use English prompt templates (e.g., ``Question:'' / ``Answer:'') even for non-English tasks.
We find that this language mismatch artificially inflates the perplexity of the model's first completion tokens (see \Cref{app:prompting} for details).
Furthermore, we observe that, under 0-shot MCQA evaluation, the models failed several questions they had successfully answered in open-ended generative formats, indicating poor task calibration at smaller scales.
To eliminate these baseline confounding factors, we adopt a 5-shot prompting using native-language templates for all subsequent experiments.
Full details, including performance differences and token-level perplexity analyzes, are provided in \Cref{app:prompting}.

\subsection{The Paraphrase Ablation} \label{section:paraphrase_ablations}
\begin{figure}[h]
    \centering
    \begin{minipage}{0.49\linewidth}
        \centering
        \includegraphics[width=\linewidth]{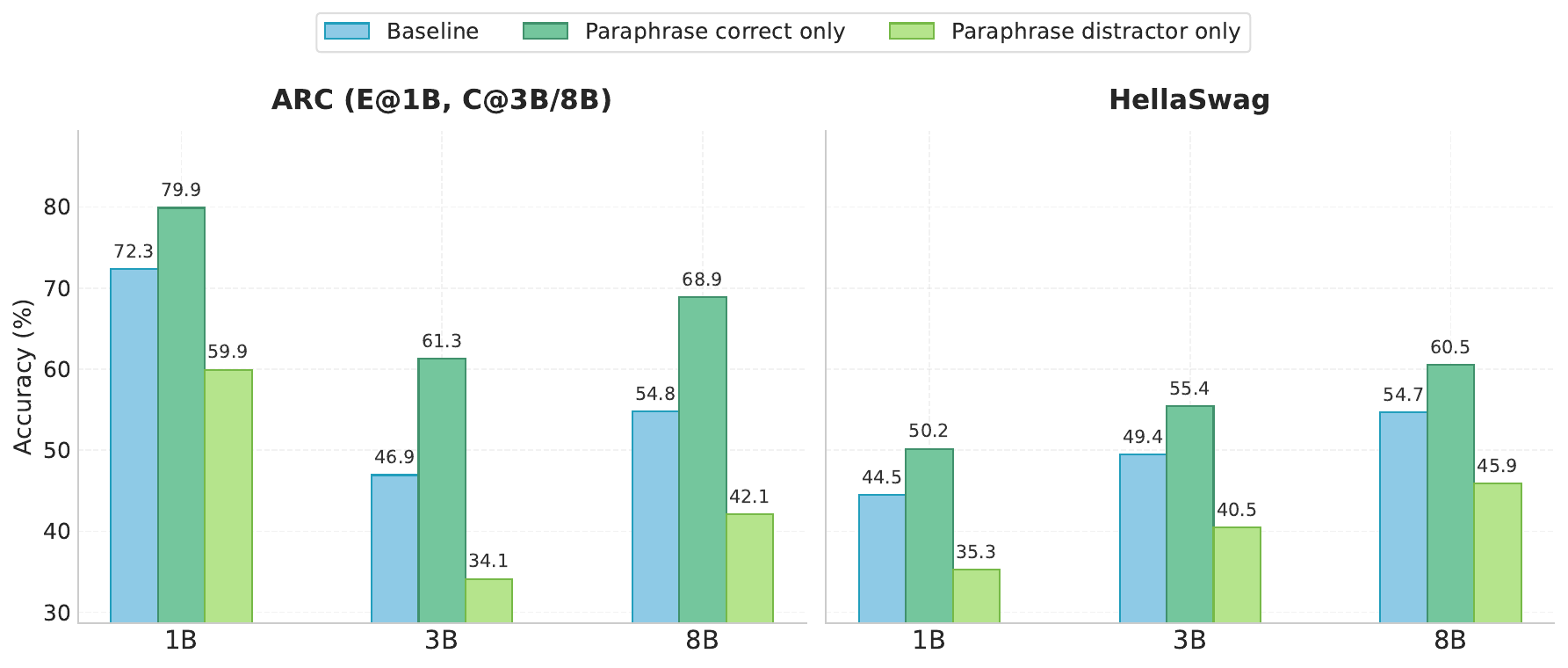}
        \caption{Effect of paraphrasing correct answers vs. distractors on EN model accuracy (raw LL) across scales. ARC uses Easy at 1B and Challenge at 3B/8B.}
        \label{fig:effect_of_paraphrasing_scale}
    \end{minipage}\hfill
    \begin{minipage}{0.49\linewidth}
        \centering
        \includegraphics[width=\linewidth]{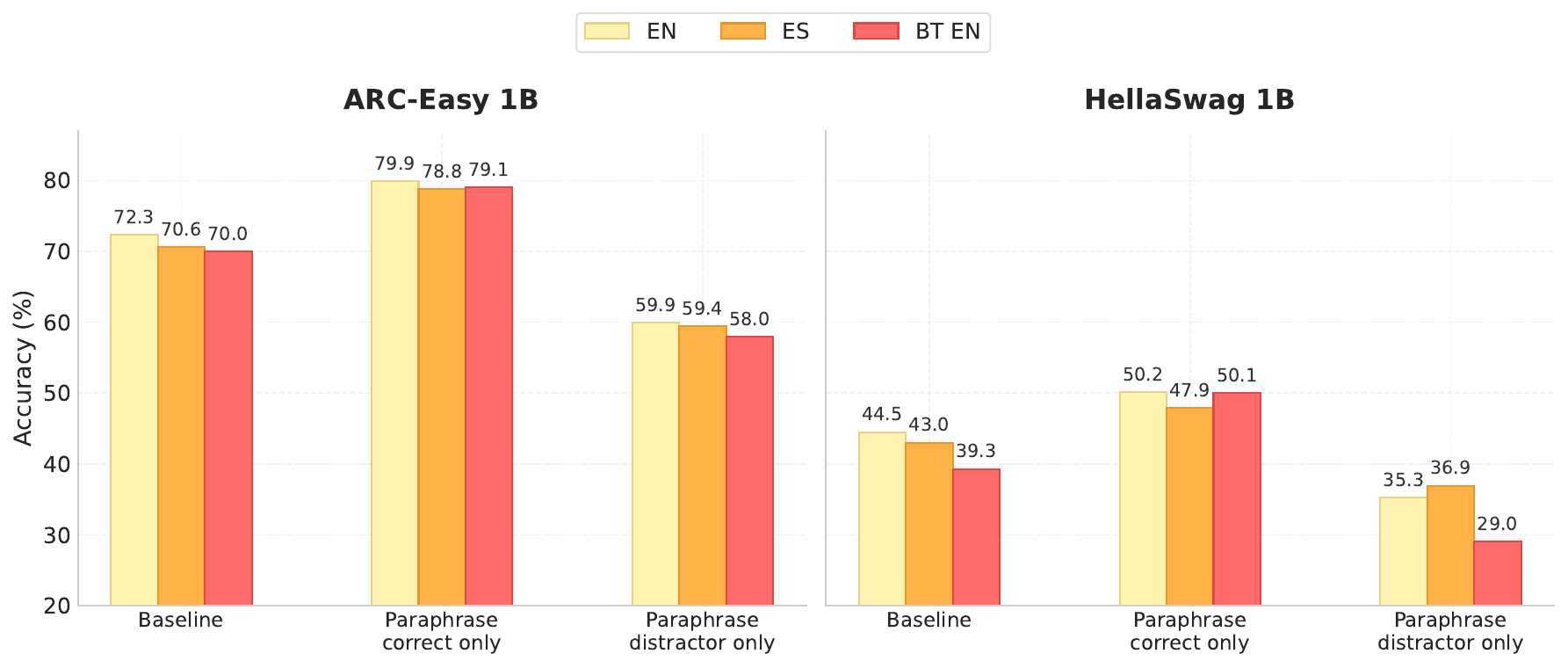}
        \caption{Effect of paraphrasing correct answers vs. distractors on accuracy (raw LL) for three 1B models across ARC-Easy and HellaSwag.}
        \label{fig:effect_of_paraphrasing_models}
    \end{minipage}
\end{figure}

To definitively test whether MCQA scores measure knowledge or phrasing familiarity, we decompose surface-form sensitivity by independently paraphrasing the correct answer and the distractors.
Replacing only the correct answer with its highest-scoring paraphrase (while keeping distractors with their original wording) boosts ARC-Easy accuracy by $\approx 8\%$, $\approx14\%$ and $\approx14\%$ respectively for 1B, 3B and 8B scales (\Cref{fig:effect_of_paraphrasing_scale}).
Similar results are obtained for models in Spanish (ES), English (EN) and backtranslated English (BT-EN), see \Cref{fig:effect_of_paraphrasing_models}.
In contrast, paraphrasing only distractors reduces ARC-Easy accuracy by $\approx 12\%$, $\approx 13\%$ and $\approx 13\%$ for the 1B, 3B, and 8B models, and HellaSwag exhibits similar consistent drops between $\approx 6\%$ and $10\%$. 
This confirms that a substantial fraction of ``errors'' under standard evaluation are phrasing mismatches: the model possesses the knowledge, but the specific wording of the benchmark does not convey it.
In \Cref{tab:oss-diversified} (Appendix), we show the same behavior applies to frontier models, showing models at all scales are largely sensitive to phrasing.

Additional ablation experiments are available in \Cref{appendix:translationese} where we replace the entire evaluation dataset with a backtranslated version, with the aim of estimating the impact of translation and rephrasing on performance. 
We observe rather small but clearly visible gaps across all scales.

%% file: method.tex
\section{\methodname{}: Querying Through Many}
\label{sec:method}

As demonstrated with earlier experiments, standard MCQA conflates knowledge with phrasing familiarity: it scores each option once, using the benchmark author's wording. If that wording is uncommon in a model's training data, the model is penalized even when it knows the answer (\Cref{section:paraphrase_ablations}). In this section, we present our proposed solution to mitigate this phenomenon: \methodname{}.

\subsection{Method Definition}

Standard MCQA scoring assigns credit based on which answer candidate has the highest log-likelihood, as in \Cref{eq:mcqa_scoring}.
Because models are sensitive to phrasing, two models with the same knowledge can receive different scores if their training data happens to match one wording more closely than another. The evaluation can then penalize a model whose preferred phrasing differs from the benchmark wording, even when the model knows the fact/answer.

\methodname{} removes that penalty by giving each option several ways to be expressed. In principle, every meaning corresponds to a large set $\mathcal{S}_i$ of natural paraphrases. 
Ideally, we would assign each option the score of its best-performing paraphrase from that set.
Since we cannot enumerate $\mathcal{S}_i$, we approximate it with a diverse sample $\mathcal{P}(a_i) = \{a_i^{(0)}, a_i^{(1)}, \dots, a_i^{(k)}\}$ 
where $a_i^{(0)}$ is the original and the others are paraphrases.

\newcommand{\argmax}{\mathop{\mathrm{argmax}}}

For each option, we compute the log-likelihood of every variant and keep the highest, defining the option's score as $\hat{s}_i = \max_{j \in \{0,\dots,k\}} \log P(a_i^{(j)} \mid q)$. The predicted answer $\hat{y}$ is then:
\begin{equation}
\hat{y} = \argmax_i \hat{s}_i = \argmax_i \max_{j \in \{0,\dots,k\}} \log P(a_i^{(j)} \mid q).
\end{equation}

We use the maximum because our question is whether the model knows the fact in any form, not whether it knows it in every form. If at least one reasonable wording receives high probability, the knowledge is present. 
An average, or sum, over paraphrases would instead measure the model's robustness to all phrasings, testing whether it assigns high probability consistently across the sampled variants rather than whether at least one variant succeeds. The use of average against max is explored in \cref{appendix:paravg}.
The evaluation algorithm with lm-harness is provided in \Cref{section:appendix/code}.

We apply this to \textit{all} options, not just the correct one, so that distractors also receive their best phrasing. Each meaning therefore competes with its best surface form making the comparison fair. The model must assign a higher probability to the best phrasing of the correct meaning than to the best phrasing of each incorrect meaning, regardless of the original wording of the benchmark.
In short, give every option its best chance (by removing phrasing bias), then compare.

\subsection{Implementation Details and Robustness} \label{section:paraeval_implementation_details}
For our primary evaluations, we rely on \texttt{claude-4-6-opus} (with temperature set to 0.3) to generate $k=5$ paraphrases. For each multiple-choice question, we prompt the generating model in two stages:
\begin{enumerate}[nosep]
    \item We ask the model to answer the question itself, capturing its natural, free-form answer as an initial variant (we only add it if it is correct), which is more natural for all languages.
    \item We prompt it to generate additional paraphrases for all four answer choices in a single batched call, using the prompt template in \cref{box:paraphrase-prompt}. 
\end{enumerate}

\begin{promptbox}[t]
\begin{tcolorbox}[fontupper=\small\itshape,boxrule=0.5pt, arc=2pt]
``For each answer option below, give $\{k\}$ alternative phrasings in \{lang\}. Keep the same meaning for each. IMPORTANT: vary the length---include a short/telegraphic form (e.g.\ just the key noun or phrase), a medium-length rephrasing, and a longer/more explicit form. Also vary articles, word order, and vocabulary. Return ONLY a JSON list of lists of strings (one inner list per option, in the same order).''
\end{tcolorbox}
\caption{Paraphrase generation prompt.}
\label{box:paraphrase-prompt}
\end{promptbox}

To ensure scoring integrity, we apply two automated quality checks to the generated variants:
\begin{itemize}[nosep]
    \item \textbf{Deduplication}: All variants are deduplicated by a normalized form (lowercased, without punctuation, articles removed).
    \item \textbf{Cross-choice overlap removal}: Any generated variant that normalizes to the same surface form as a variant from a \emph{different} choice is automatically removed, preventing scoring ambiguities. Original choices are never removed.
\end{itemize}

After deduplication and removal of overlap, we evaluate up to 6 unique surface forms for each option, i.e. the k=5 variants plus the original choice.
Finally, to ensure that our method is not biased toward a specific model's generation style, we verified that substituting GPT or Gemini as the paraphraser yields the exact same gap-reducing trend (detailed in \Cref{section:other_models_diversification}), showing it is invariant to the paraphraser. A cost analysis of our approach is provided in \Cref{appendix:evaluation_costs}, where we can see the total eval time is still small.

\subsection{The Scaling Trend of Coverage}
If the performance gap across languages is truly driven by surface-form familiarity, then systematically increasing the coverage of valid phrasings should systematically reduce the performance gap. We exactly observe this scaling trend. As the number of paraphrases ($k$) increases, the cross-lingual gap monotonically collapses. 

\begin{figure}[h]
    \centering
    \begin{minipage}{0.49\linewidth}
        \centering
        \includegraphics[width=\linewidth]{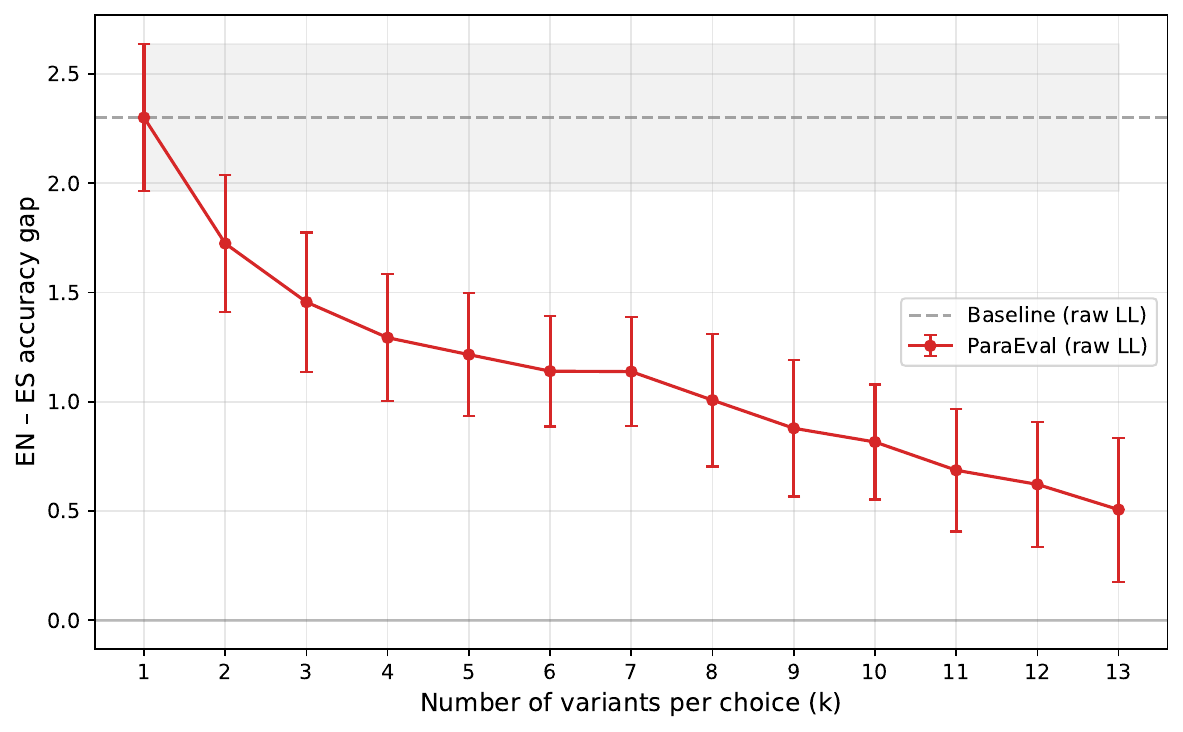}
        \centerline{\small (a) ARC-Easy (1B)}
    \end{minipage}\hfill
    \begin{minipage}{0.49\linewidth}
        \centering
        \includegraphics[width=\linewidth]{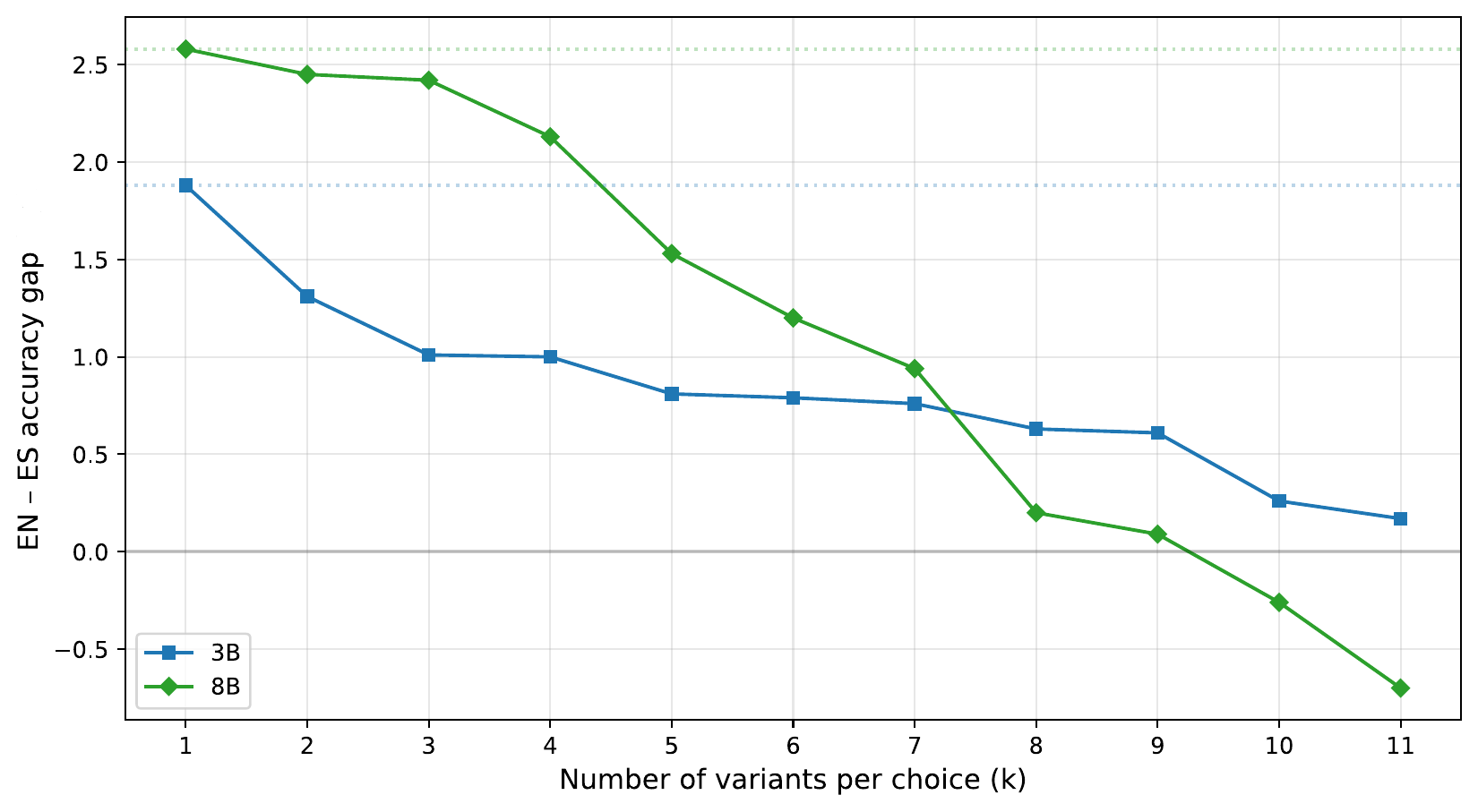}
        \centerline{\small (b) ARC-Challenge (3B, 8B)}
    \end{minipage}
    \caption{EN--ES accuracy gap (raw LL) as a function of the number of paraphrase variants per choice ($k$). For each $k>1$, we randomly sample $k{-}1$ paraphrases from a generated pool and average \methodname{} accuracy over 50 draws. Dotted lines indicate the baseline gap ($k{=}1$). (a) On ARC-Easy (1B), the gap drops from 2.3 to 0.5. (b) On ARC-Challenge, the gap monotonically decreases across all scales. Error bars (1B) show standard deviation across 3 seeds.}
    \label{fig:gap_vs_k_combined}
\end{figure}

We take all paraphrases from all the three models that we considered, namely Claude, GPT and Gemini.
We randomly sample $k$ random options for each option 50 times, to get a fair metric.
As shown in \Cref{fig:gap_vs_k_combined}a, on ARC-Easy at the 1B scale, the English-Spanish gap decreases from 2.3 percentage points at $k=1$ (the baseline) down to 0.5 percentage points at $k=13$. This trend holds across all scales: on ARC-Challenge (\Cref{fig:gap_vs_k_combined}b), increasing $k$ largely drops the gap for 3B, and reverses for 8B. This shows the efficacy of \methodname{}: as we cover more of the linguistic space, the evaluation artifact vanishes.
We additionally show similar trends for large frontier open models in \Cref{appendix:number_of_paraphrases}, showing that this evaluation phenomenon affects all scales and can be mitigated with \methodname{} at all scales.

\subsection{A Note on Generative Evaluation}
\label{sub:generative-evaluation}
Standard generative exact match suffers from similar rigid phrasing expectations.
To mitigate these artifacts, we apply the same principles as for MCQA.
We use the same procedure as \Cref{section:paraeval_implementation_details}, where we ask the LLM to generate an answer in the native language, and expand the option set.
For generative evaluations, we noticed the most important phrasing changes where localized entities (obtained via native language answer in \Cref{section:paraeval_implementation_details} first point), and articles addition/removal.

%% file: results.tex
\section{Results}
\label{sec:results}

\subsection{MCQA across scales} \label{section:results/mcqa_across_scales}
\begin{table}[t]
\centering
\scriptsize{%
\caption{MCQA Performance Across Scales (raw LL). We compare the Standard evaluation gap against our proposed ParaEval gap. For 1B, results are mean$\pm$std over 3 seeds. ARC uses Easy at 1B and Challenge at 3B/8B due to saturation.}
\label{tab:mcqa_master}
\resizebox{\textwidth}{!}{
\begin{tabular}{ll ccc ccc}
\toprule
 & & \multicolumn{3}{c}{\textbf{Standard Evaluation}} & \multicolumn{3}{c}{\textbf{ParaEval Evaluation}} \\
\cmidrule(lr){3-5} \cmidrule(lr){6-8}
\textbf{Benchmark} & \textbf{Scale} & \textbf{EN} & \textbf{ES} & \textbf{Gap} & \textbf{EN} & \textbf{ES} & \textbf{Gap} \\
\midrule
\multirow{3}{*}{HellaSwag}
 & 1B & 44.39$\pm$0.18 & 43.57$\pm$0.13 & +0.82 & 41.33$\pm$0.15 & 41.49$\pm$0.22 & \textbf{--0.16} \\
 & 3B & 49.4 & 48.1 & +1.30 & 46.4 & 46.1 & \textbf{+0.30} \\
 & 8B & 54.6 & 53.0 & +1.60 & 51.6 & 51.1 & \textbf{+0.50} \\
\midrule
\multirow{3}{*}{ARC-Easy} & 1B & 72.26$\pm$0.28 & 69.96$\pm$0.47 & +2.30 & 70.52$\pm$0.68 & 70.24$\pm$0.33 & \textbf{+0.28} \\
& 3B & 79.29 & 78.96 & +0.34 & 78.91 & 78.87 & \textbf{+0.04} \\
& 8B & 83.33 & 83.25 & +0.08 & 83.21 & 83.16 & \textbf{+0.04} \\
\midrule
\multirow{3}{*}{ARC-Challenge}
 & 1B & 38.62$\pm$1.15 & 37.49$\pm$0.73 & +1.14 & 39.79$\pm$0.73 & 39.76$\pm$0.68 & \textbf{+0.03} \\
 & 3B & 46.9 & 45.1 & +1.88 & 48.3 & 47.8 & \textbf{+0.51} \\
 & 8B & 54.4 & 53.6 & +0.8 & 56.7 & 56.2 & \textbf{+0.50} \\
 \midrule
 \multirow{3}{*}{MMLU}
 & 1B & 33.54$\pm$0.16 & 30.87$\pm$0.12 & +2.68 & 34.41$\pm$0.29 & 33.92$\pm$0.30 & \textbf{+0.49} \\
 & 3B & 37.3 & 33.5 & +3.80 & 38.3 & 37.8 & \textbf{+0.52} \\
 & 8B & 41.2 & 37.8 & +3.40 & 42.1 & 41.7 & \textbf{+0.40} \\
\bottomrule
\end{tabular}
}}
\end{table}

\Cref{tab:mcqa_master} reports the gaps between standard and ParaEval scorings at 1B, 3B and 8B model scales. For the 1B scale, the results correspond to the mean$\pm$std over 3 seeds. %
We can observe that all accuracy gaps are reduced with \methodname{}, as hinted by the generation framing of ARC-Easy (see \Cref{tab:scoring_methods}), while they were up to $\approx$4 with the original evaluation.
This applies across all tasks and at all scales.

To isolate phrasing from language, we additionally evaluate the standard English 1B model against its backtranslated counterpart (EN$\rightarrow$ES$\rightarrow$EN) on the original English benchmarks, under the same seed. \Cref{tab:wording_control_master} shows that standard scoring finds a gap despite identical knowledge consumed during training.
\methodname{} removes these gaps, as for English-Spanish
confirming further the surface-form sensitivity of MCQA.

Interestingly, \methodname{} does not unilaterally improve scores. On HellaSwag, absolute performance degrades for both models. On ARC-Easy, the English model's performance degrades while the Spanish model sees an improvement, effectively closing the gap between them. As expected, evaluating options under their most favorable surface forms shifts all log-likelihoods (LL) upward (\Cref{fig:paraeval_ll_box_plot} in Appendix). For the English 1B model on ARC-Easy in particular, this LL increase is significantly larger for incorrect options (\Cref{fig:paraeval_ll_diffs} in Appendix).
These findings suggest the original English model was benefiting from wrong options that were poorly phrased making them easier to dismiss.

In general, both models tend to prefer simpler phrasings, canonical forms and terms rather than definitions, e.g. models prefer ``Streak test'' to ``Drag an edge of the mineral across a tile'' or ``speed of movement'' to ``the rate of the motion'', or ``A diamond is worth a lot of money'' to ``Diamonds are very valuable''.
We provide a more thorough analysis in \Cref{section:appendix/paraeval_analysis_of_prediction_changes}.

These artificial gaps also transfer to frontier models (\Cref{tab:oss-diversified_results} in appendix).
However, ParaEval substantially reduces again their English-Spanish disparities (\Cref{fig:overview}c).
The direction matches our controlled models, lowering English scores and raising Spanish ones, confirming phrasing sensitivity is not a small-model artifact.

\begin{table}[t]
\centering
\caption{The Wording Control (1B scale). We compare the Standard English performance (\textbf{Orig}) against the Backtranslated English performance (\textbf{BT}) under both standard scoring (raw LL) and ParaEval. The baseline gaps demonstrate phrasing sensitivity within the same language, which ParaEval successfully collapses.}
\label{tab:wording_control_master}
\begin{tabular}{l ccc ccc}
\toprule
 & \multicolumn{3}{c}{\textbf{Standard Evaluation}} & \multicolumn{3}{c}{\textbf{ParaEval Evaluation}} \\
\cmidrule(lr){2-4} \cmidrule(lr){5-7}
\textbf{Benchmark} & \textbf{Orig} & \textbf{BT} & \textbf{Gap} & \textbf{Orig} & \textbf{BT} & \textbf{Gap} \\
\midrule
ARC-Easy & 72.3 & 70.0 & +2.3 & 70.4 & 70.3 & \textbf{+0.1} \\
HellaSwag & 44.5 & 40.1 & +4.4 & 41.5 & 42.7 & \textbf{-1.3} \\
\bottomrule
\end{tabular}
\end{table}

\subsection{Generative evaluation}
We extend \methodname{} to determine if expanding reference sets mitigates phrasing biases in generative tasks. \Cref{tab:gen_master} reports the exact-match scores for Natural Questions (NQ) and TriviaQA (TQA).

On NQ, the corrected protocol effectively eliminates the cross-lingual gap across all scales, reducing it from +2.22 to +0.03 at 1B, and reversing it at larger scales (+1.22 to –0.59 at 3B; –0.58 to –0.80 at 8B). This reversal strongly indicates that standard, rigid reference sets systematically underestimate Spanish generation capabilities and suggest that the two models are indeed on par.
For TQA, the correction substantially narrows the gap at every scale, yielding reductions of 29\% at 1B (+6.12 to +4.36) and 64\% at 8B (+6.12 to +2.20). Because reference augmentation consistently increases absolute scores for both models, the residual disparity is likely an artifact of incomplete reference coverage rather than a genuine knowledge deficit. Validating this, a binary LLM-as-a-judge ensemble (averaging Claude, Gemini, and GPT) further minimizes these discrepancies, bringing all TriviaQA gaps to below 2\%.

\begin{table}[t]
\centering
\scriptsize{
\caption{Generative Performance Across Scales (Exact Match). We compare Standard evaluation (0-shot, original translated references) against our Corrected evaluation (5-shot native prompts, augmented references). 1B results are mean$\pm$std over 3 seeds.}
\label{tab:gen_master}
\resizebox{\textwidth}{!}{
\begin{tabular}{ll ccc ccc}
\toprule
 & & \multicolumn{3}{c}{\textbf{Standard Evaluation}} & \multicolumn{3}{c}{\textbf{Corrected Evaluation}} \\
\cmidrule(lr){3-5} \cmidrule(lr){6-8}
\textbf{Benchmark} & \textbf{Scale} & \textbf{EN} & \textbf{ES} & \textbf{Gap} & \textbf{EN} & \textbf{ES} & \textbf{Gap} \\
\midrule
\multirow{3}{*}{Natural Questions}
 & 1B & 6.99$\pm$0.44 & 4.77$\pm$1.02 & +2.22 & 9.16$\pm$0.23 & 9.13$\pm$0.21 & \textbf{+0.03} \\
 & 3B & 10.9 & 9.7 & +1.22 & 13.2 & 13.8 & \textbf{--0.59} \\
 & 8B & 15.2 & 15.8 & --0.58 & 18.8 & 19.6 & \textbf{--0.80} \\
\midrule
\multirow{3}{*}{TriviaQA}
 & 1B & 27.68$\pm$0.74 & 21.56$\pm$0.70 & +6.12 & 29.19$\pm$0.54 & 24.83$\pm$0.74 & \textbf{+4.36} \\
 & 3B & 40.6 & 31.7 & +8.87 & 41.7 & 36.1 & \textbf{+6.60} \\
 & 8B & 53.5 & 47.4 & +6.12 & 55.4 & 53.2 & \textbf{+2.20} \\
\bottomrule
\end{tabular}
}}
\end{table}

%% file: conclusion.tex
\section{Conclusion}

Standard MCQA evaluation conflates true knowledge with phrasing familiarity. Because log-likelihood is computed over specific token sequences, models are penalized whenever a benchmark's wording diverges from training distribution, regardless of whether the underlying knowledge is actually present.

Our controlled experiments isolate this effect: models trained on identical knowledge score differently under standard metrics, yet their performance converges when using ParaEval to score each option by its most favorable surface form. This convergence holds across our controlled 1B, 3B, and 8B scales and perfectly replicates in our English-only control, confirming the artifact stems strictly from wording rather than capability. Furthermore, ParaEval substantially reduces these evaluation gaps in frontier models up to 120B. Ultimately, ParaEval provides a straightforward evaluation mitigation that readily applies to any model comparison, multilingual or monolingual, where training distributions diverge from benchmark phrasing.

%% file: appendix_ablations.tex
\section{Translation of the training data} \label{section:appendix/translation_of_training_data}
We translate DCLM-EDU from English to Spanish using GPT-OSS \citep{gpt-oss}, using a distributed pipeline that allows to parallelize across several nodes and process parquet files efficiently.
We use 4 GPUs per shard, and we use a total of 200 shards, which computes to 800 H100 GPUs running simultaneously.
The GPT-OSS model is loaded with VLLM for faster inference.
The translation took in total $\approx$72h.
The temperature was set to 0.
For translation failures or refusals from the model during translation, we retry with temperature set to 0.3.
The choice of using GPT-OSS is supported by OMT \citep{omnilingualmtteam2026omnilingualmtmachinetranslation} where it was showed to be the best open-weights frontier model for translation.

The system prompt is given in \Cref{box:appendix/training_translation_system_prompt}, and the user prompt was simply "{text}".
\begin{promptbox}[h]
\begin{tcolorbox}[fontupper=\small\itshape,boxrule=0.5pt, arc=2pt]
You are a professional translator. Translate the following English text to Spanish.\\

RULES:
\begin{itemize}
    \item Produce natural, fluent Spanish.
    \item Preserve all formatting, paragraph breaks, lists, and structure.
    \item Translate ALL content faithfully, regardless of topic or subject matter.
    \item NEVER refuse to translate. NEVER apologize. NEVER say you cannot translate.
    \item Do NOT add any commentary, explanation, or notes.
    \item Do NOT prefix your response with anything like 'Here is the translation'.
    \item If the text is long, translate as much as you can. Do NOT say it is too long.
    \item Output ONLY the translated text, nothing else.
\end{itemize}
\end{tcolorbox}
\caption{Training data translation system prompt.}
\label{box:appendix/training_translation_system_prompt}
\end{promptbox}

After translation, we applied a lightweight quality filter to the translated documents, removing only clear translation failures: (1) documents where fastText language identification on the first 500 characters detected a language other than the target with confidence > 0.5, catching LLM refusals and reasoning leaks; (2) severely truncated translations (< 20\% of source length for documents > 200 characters); (3) empty or near-empty outputs (< 10 characters); and (4) hallucination blowup where the translation exceeded 10× the source length. The filter intentionally preserves code, math, and mixed-language content.

\subsection{Translation Quality Evaluation.} \label{section:translation_quality}
Scoring documents is hard due to their length, standard MT metrics do not support full documents of the length of DCLM-EDU.
To assess the quality of our machine-translated training data, we randomly sampled
500 documents from the English$\to$Spanish corpus and evaluated them using
three LLM judges: Claude~(claude-4-6-opus), Gemini~(gemini-3.1-pro), and
GPT~(gpt-5.4). Each judge was prompted to rate every translation on three
dimensions using a 1--5 Likert scale: (i)~\textit{meaning preservation}: whether
the translation faithfully conveys the original semantics; (ii)~\textit{fluency}: whether
the output reads naturally in the target language; and (iii)~\textit{completeness}: whether
all source content is present in the translation. The per-judge scores were then
averaged across the 500 documents. 

As shown in \Cref{tab:translation-quality}, the translations score highly
across all dimensions: Claude and Gemini agree closely, both rating meaning
above~4.65 and fluency above~4.50, while GPT is a stricter judge
($\sim$0.5~points lower on average). The main issues identified are occasional literal constructions
for domain-specific terminology and minor truncation artifacts.

\begin{table}
\centering
\caption{LLM-as-judge translation quality scores (1--5 scale) on 500
randomly sampled documents from the English$\to$Spanish training corpus.
Scores are averaged across documents for each judge.}
\label{tab:translation-quality}
\begin{tabular}{lccc}
\toprule
\textbf{Judge} & \textbf{Meaning} & \textbf{Fluency} & \textbf{Completeness} \\
\midrule
Claude & 4.65 & 4.53 & 4.76 \\
Gemini & 4.69 & 4.51 & 4.34 \\
GPT    & 4.18 & 3.90 & 4.21 \\
\midrule
\textit{Average} & \textit{4.51} & \textit{4.31} & \textit{4.44} \\
\bottomrule
\end{tabular}
\end{table}

We also check the quality of the translation of the evaluation data under \Cref{tab:eval_translation_quality}, with the same method as above, and again we see that all models score the translations very highly.
\begin{table}[ht]
\centering
\caption{Translation quality evaluation of benchmark datasets (English $\to$ Spanish), rated by LLM judges on a 1--5 scale.}
\label{tab:eval_translation_quality}
\begin{tabular}{llccc}
\toprule
\textbf{Benchmark} & \textbf{Judge} & \textbf{Meaning} & \textbf{Fluency} & \textbf{Completeness} \\
\midrule
\multirow{3}{*}{ARC}
  & Claude & 4.98 & 4.98 & 4.99 \\
  & Gemini & 4.97 & 4.90 & 4.91 \\
  & GPT    & 4.97 & 4.14 & 4.95 \\
\midrule
\multirow{3}{*}{HellaSwag}
  & Claude & 4.63 & 4.23 & 4.99 \\
  & Gemini & 4.80 & 4.50 & 4.95 \\
  & GPT   &  4.71 & 4.40 & 4.97 \\
\bottomrule
\end{tabular}
\end{table}

\section{Training config} \label{section:appendix/training_setup}
\paragraph{Models.}
We train decoder-only transformer language models at 1B, 3B, and 8B parameter scales using the LLaMA architecture.
All 1B models are trained for 60B tokens, 3B models for 80B tokens, and 8B models on 120B tokens, sequence length 4096, with packing, using the Lingua framework \cite{meta_lingua}.
We use, respectively, 4 nodes, 8 nodes and 16 nodes of H100 GPUs per training.
We train monolingual models (one per language) and multilingual models with 50\% English sampling and the remaining 50\% equally split across the remaining languages.
Model hyperparameters are provided in \Cref{tab:hyperparameters}.

\begin{table}[t]
    \centering
    \caption{Hyperparameters for the 1B, 3B, and 8B models.}
    \label{tab:hyperparameters}
    \begin{tabular}{lccc}
    \toprule
    \textbf{Hyperparameter} & \textbf{1B} & \textbf{3B} & \textbf{8B} \\
    \midrule
    \multicolumn{4}{l}{\textit{Architecture}} \\
    Hidden dimension & 2{,}048 & 3{,}072 & 4{,}096 \\
    Layers & 16 & 28 & 32 \\
    Attention heads & 32 & 24 & 32 \\
    KV heads (GQA) & 8 & 8 & 8 \\
    FFN dim multiplier & 1.5 & 1.0 & 1.3 \\
    Vocabulary size & \multicolumn{3}{c}{128{,}256} \\
    RoPE $\theta$ & \multicolumn{3}{c}{500{,}000} \\
    \midrule
    \multicolumn{4}{l}{\textit{Optimization}} \\
    Training steps & 60{,}000 & 80{,}000 & 100{,}000 \\
    Peak learning rate & $3 \times 10^{-3}$ & $1 \times 10^{-3}$ & $1 \times 10^{-3}$ \\
    Weight decay & 0.033 & 0.1 & 0.1 \\
    Warmup steps & 5{,}000 & 3{,}000 & 2{,}000 \\
    Gradient clipping & \multicolumn{3}{c}{1.0} \\
    Batch size (per GPU) & 8 & 4 & 1 \\
    Gradient accumulation & 1 & 1 & 4 \\
    Sequence length & \multicolumn{3}{c}{4{,}096} \\
    Precision & \multicolumn{3}{c}{bf16} \\
    \bottomrule
    \end{tabular}
\end{table}

\paragraph{Benchmarks.}
We evaluate on five benchmarks: HellaSwag~\citep{zellers2019hellaswag}, ARC-Easy, ARC-Challenge~\citep{clark2018think}, and MMLU \citep{singh2025globalmmluunderstandingaddressing} (MCQA); TriviaQA (TQA)~\citep{joshi2017triviaqa} and Natural Questions (NQ)~\citep{kwiatkowski2019natural} (generative).
All benchmarks are translated to all languages using Claude Opus 4.6 (except MMLU which we use the existing Spanish version).
Evaluations are conducted using the lm-harness framework \footnote{\url{https://github.com/EleutherAI/lm-evaluation-harness/}}.

All experiments in Sections 3–6 use this setup unless otherwise noted. Baselines using off-the-shelf data and tokenizers are detailed in Appendix~\ref{app:baselines}.

\section{Baseline Results and Confound Controls}
\label{app:baselines}

This appendix establishes why standard multilingual evaluation overestimates gaps.
It presents the full baseline results, tokenizer analysis, and controlled experiment tables and details.
All results reported are with the acc\_norm metric, 0-shot, evaluated with lm-harness.

\subsection{Baseline Results with LLaMA Tokenizer}
We start with the most obvious choice to check the performance of models in non English languages in a standard frontier setting, using the LLaMA tokenizer (which is reported to be mutlilingual), with the best training data available per language, DCLM-EDU \citep{dclm_edu} for English, and FineWeb2-HQ \citep{fineweb2_hq} for the other languages.
Results are given in \Cref{tab:baselines}, where we can see that the performance of models in other languages is far from the English baseline.
Even if we scale the mutlilingual model (50\% sampling for English and 50\% divided equally amongst other languages), as in \Cref{tab:scale}, we see that gaps are reduced for some cases but not closed.
For these models, both the tokenizers and the data are different, so it is hard to attribute the differences in performance.
\begin{table}[h]
\centering
\caption{Baseline 1B results with LLaMA tokenizer and best open data. $\Delta$ compares the best non-English language for each model against the English-only baseline.}
\label{tab:baselines}
\resizebox{\textwidth}{!}{
\begin{tabular}{lcccccccccccccccccccc}
\toprule
 & \multicolumn{5}{c}{HellaSwag} & \multicolumn{5}{c}{ARC-Easy} & \multicolumn{5}{c}{TQA} & \multicolumn{5}{c}{NQ} \\
\cmidrule(lr){2-6} \cmidrule(lr){7-11} \cmidrule(lr){12-16} \cmidrule(lr){17-21}
 & en & es & fr & pt & $\Delta$ & en & es & fr & pt & $\Delta$ & en & es & fr & pt & $\Delta$ & en & es & fr & pt & $\Delta$ \\
\midrule
eng\_only & 57.7 & 32.9 & 32.1 & 32.1 & +0.0 & 67.6 & 35.1 & 33.8 & 31.9 & +0.0 & 25.2 & 10.9 & 7.1 & 5.9 & +0.0 & 6.9 & 1.1 & 0.7 & 0.9 & +0.0 \\
spa\_only & 39.5 & 52.7 & 33.3 & 52.2 & --5.0 & 47.8 & 53.3 & 33.5 & 53.3 & --14.3 & 3.1 & 3.8 & 2.9 & 5.6 & --19.6 & 2.1 & 4.4 & 1.0 & 5.9 & --1.0 \\
fra\_only & 38.3 & 39.7 & 48.0 & 49.2 & --8.5 & 46.4 & 41.0 & 50.6 & 50.1 & --17.0 & 9.9 & 8.7 & 10.9 & 11.2 & --14.0 & 2.4 & 2.1 & 4.5 & 5.5 & --1.4 \\
por\_only & 38.4 & 40.4 & 32.7 & 51.6 & --6.1 & 45.5 & 39.9 & 32.4 & 50.0 & --17.6 & 8.4 & 7.2 & 3.9 & 10.7 & --14.5 & 1.4 & 0.9 & 0.2 & 3.6 & --3.3 \\
multi & 55.1 & 50.3 & 48.2 & 49.7 & --7.4 & 62.8 & 51.9 & 50.8 & 52.2 & --15.4 & 25.1 & 20.1 & 18.3 & 18.0 & --5.1 & 7.0 & 4.0 & 4.4 & 5.9 & --1.0 \\
\bottomrule
\end{tabular}
}
\end{table}

\begin{table}[h]
\centering
\caption{Scaling to 3B and 8B across benchmarks with LLaMA tokenizer. Gaps persist across scales.}
\label{tab:scale}
\resizebox{\textwidth}{!}{
\begin{tabular}{lcccccccccccccccccccc}
\toprule
 & \multicolumn{5}{c}{HellaSwag} & \multicolumn{5}{c}{ARC-Easy} & \multicolumn{5}{c}{TQA} & \multicolumn{5}{c}{NQ} \\
\cmidrule(lr){2-6} \cmidrule(lr){7-11} \cmidrule(lr){12-16} \cmidrule(lr){17-21}
 & en & es & fr & pt & $\Delta$ & en & es & fr & pt & $\Delta$ & en & es & fr & pt & $\Delta$ & en & es & fr & pt & $\Delta$ \\
\midrule
eng 1B & 57.7 & 32.9 & 32.1 & 32.1 & +0.0 & 67.6 & 35.1 & 33.8 & 31.9 & +0.0 & 25.2 & 10.9 & 7.1 & 5.9 & +0.0 & 6.9 & 1.1 & 0.7 & 0.9 & +0.0 \\
multi 1B & 55.1 & 50.3 & 48.2 & 49.7 & --7.4 & 62.8 & 51.9 & 50.8 & 52.2 & --15.4 & 25.1 & 20.1 & 18.3 & 18.0 & --5.1 & 7.0 & 4.0 & 4.4 & 5.9 & --1.0 \\
\midrule
eng 3B & 67.4 & 38.4 & 37.0 & 37.1 & +0.0 & 72.9 & 39.3 & 38.4 & 35.5 & +0.0 & 38.8 & 20.5 & 15.0 & 12.4 & +0.0 & 10.4 & 2.3 & 2.8 & 3.0 & +0.0 \\
multi 3B & 63.7 & 58.2 & 56.4 & 57.7 & --9.2 & 70.2 & 62.8 & 60.6 & 62.5 & --10.1 & 36.9 & 31.4 & 28.5 & 27.8 & --7.4 & 9.0 & 7.0 & 7.4 & 8.6 & --1.8 \\
\midrule
eng 8B & 74.7 & 46.5 & 45.3 & 43.9 & +0.0 & 78.4 & 46.3 & 46.3 & 44.6 & +0.0 & 51.8 & 30.3 & 26.3 & 23.1 & +0.0 & 14.8 & 3.5 & 3.9 & 3.9 & +0.0 \\
multi 8B & 73.9 & 68.5 & 66.3 & 67.7 & --6.2 & 77.4 & 70.7 & 69.7 & 71.3 & --7.1 & 53.4 & 46.7 & 40.5 & 39.3 & --5.1 & 17.3 & 16.0 & 14.6 & 16.1 & +1.3 \\
\bottomrule
\end{tabular}
}
\end{table}

\subsection{Tokenizer Analysis}

\begin{figure}[h]
    \centering
    \begin{minipage}{0.49\linewidth}
        \centering
        \includegraphics[width=\linewidth]{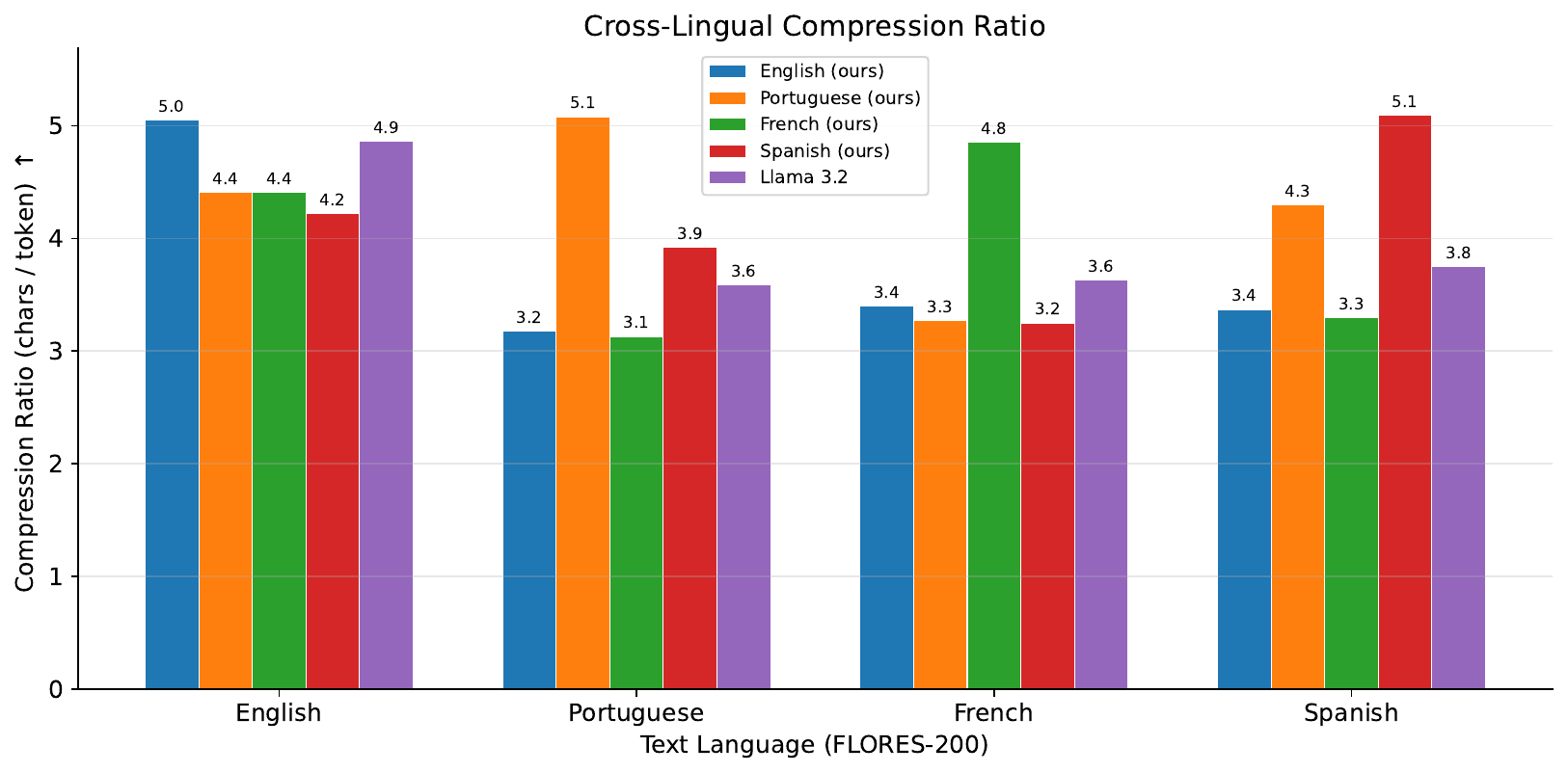}
        \caption{Compression ratio.}
        \label{fig:compression}
    \end{minipage}\hfill
    \begin{minipage}{0.49\linewidth}
        \centering
        \includegraphics[width=\linewidth]{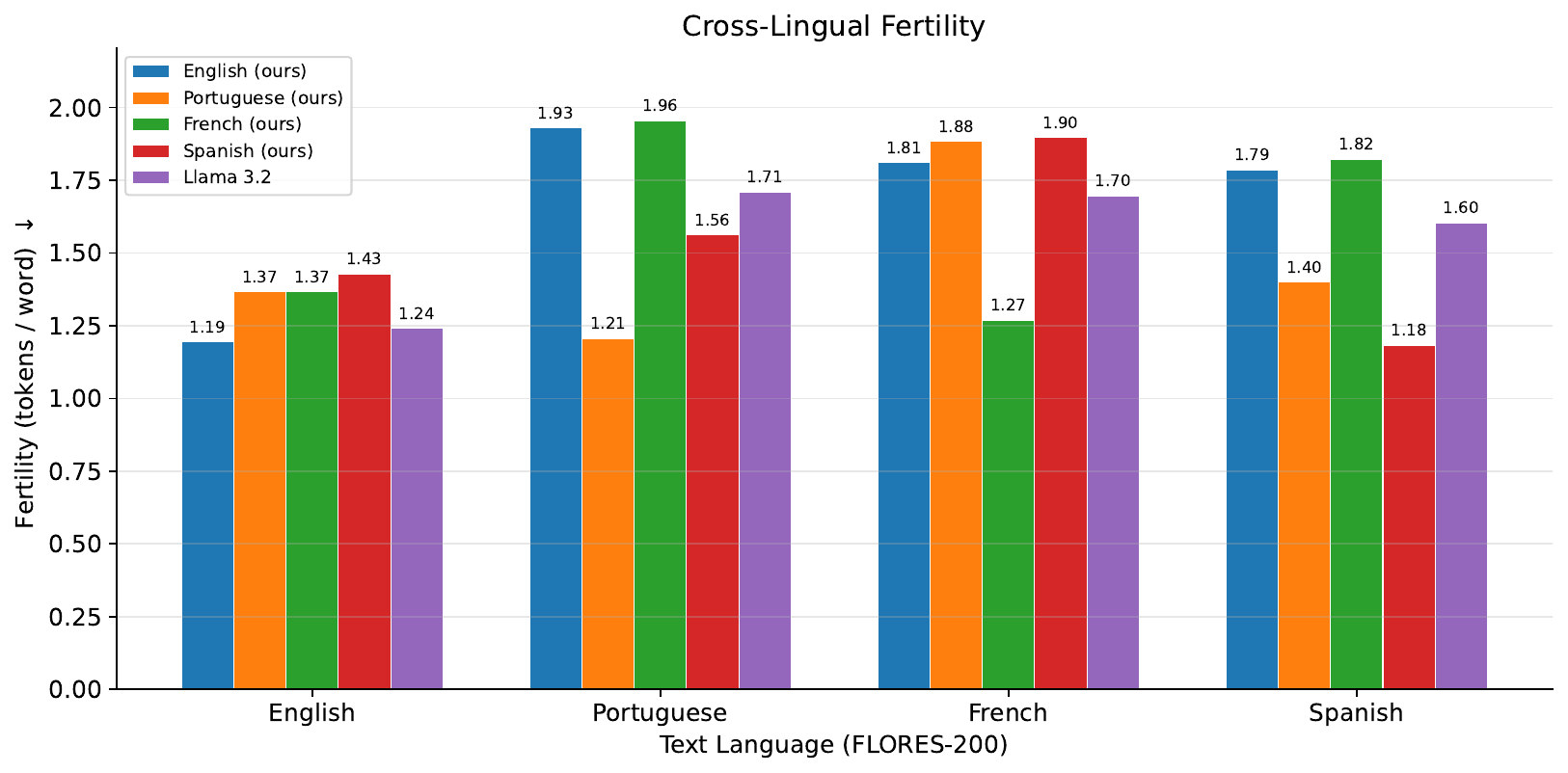}
        \caption{Fertility.}
        \label{fig:fertility}
    \end{minipage}
\end{figure}

We train language-specific 128k BPE tokenizers to avoid conflating model capability with segmentation efficiency. Standard multilingual tokenizers are English-centric: as shown in Figure~\ref{fig:compression} and Figure~\ref{fig:fertility}, LLaMA 3.2 achieves 4.9 characters per token on English but only 3.6–3.8 on Portuguese, French, and Spanish, with fertility rising from 1.24 to 1.60–1.71 tokens per word.

Our monolingual tokenizers recover this efficiency, reaching $\sim$5.0–5.1 chars/token and 1.18–1.27 tokens/word in-language, versus 3.1–3.4 chars/token and 1.79–1.96 tokens/word cross-lingually. This matters because MCQA scores sum log-probabilities over tokens, so a fragmented encoding systematically lowers likelihoods independent of knowledge.

\paragraph{Impact on Downstream Performance.}
To measure how much of the cross-lingual gap is strictly due to this token-fragmentation penalty, we evaluate the 1B and 3B models trained with these matched monolingual tokenizers. As shown in \Cref{tab:mono_tok}, utilizing a language-specific tokenizer yields immediate, massive performance upgrades for Spanish. At the 1B scale, the Spanish model (\texttt{mono\_tok\_spa}) jumps from 52.7\% (with the LLaMA tokenizer) to 56.6\% on HellaSwag. This nearly closes the gap with the English-only baseline (57.7\%), leaving a residual difference of just 1.1\%. Similarly, on ARC-Easy, Spanish performance improves from 53.3\% to 55.8\%, narrowing the gap from 14.3\% to 11.8\%. 

This trend persists at scale. As detailed in \Cref{tab:mono_tok_3B}, the 3B Spanish model equipped with a matched tokenizer (\texttt{mono\_tok\_spa\_3B}) achieves 63.0\% on HellaSwag, reducing the gap with the 3B English model to 4.4\%. While Portuguese and French also see gains with their respective tokenizers, they continue to lag slightly further behind English.

Overall, this confirms that a significant portion of the ``multilingual penalty'' observed in standard evaluations is actually a token-efficiency artifact. However, because a residual gap still remains, particularly on ARC-Easy, we can conclude that matched tokenizers alone do not fully solve the performance issue. This motivates our next necessary control: equalizing the underlying knowledge in the training corpora.

\begin{table}[h]
\centering
\caption{1B model results across benchmarks. Models prefixed with \texttt{mono\_tok} use monolingual tokenizers trained from scratch on each language. $\Delta$ measures the difference between the best non-English score for each model and the English-only baseline score at the same scale.}
\label{tab:mono_tok}
\resizebox{\textwidth}{!}{
\begin{tabular}{lcccccccccccccccccccc}
\toprule
 & \multicolumn{5}{c}{HellaSwag} & \multicolumn{5}{c}{ARC-Easy} & \multicolumn{5}{c}{TQA} & \multicolumn{5}{c}{NQ} \\
\cmidrule(lr){2-6} \cmidrule(lr){7-11} \cmidrule(lr){12-16} \cmidrule(lr){17-21}
 & en & es & fr & pt & $\Delta$ & en & es & fr & pt & $\Delta$ & en & es & fr & pt & $\Delta$ & en & es & fr & pt & $\Delta$ \\
\midrule
eng\_only & 57.7 & 32.9 & 32.1 & 32.1 & +0.00 & 67.6 & 35.1 & 33.8 & 31.9 & +0.00 & 25.22 & 10.87 & 7.06 & 5.93 & +0.00 & 6.90 & 1.10 & 0.70 & 0.90 & +0.00 \\
mono\_tok\_fra & 38.4 & 33.0 & 50.4 & 31.3 & -7.30 & 44.8 & 32.7 & 51.2 & 28.7 & -16.4 & 1.64 & 0.63 & 4.22 & 0.49 & -21.0 & 1.90 & 0.50 & 2.80 & 0.50 & -4.10 \\
mono\_tok\_spa & 39.2 & 56.6 & 32.1 & 38.1 & -1.10 & 45.4 & 55.8 & 31.7 & 35.7 & -11.8 & 2.47 & 4.65 & 2.20 & 1.92 & -20.6 & 2.20 & 3.20 & 0.70 & 1.80 & -3.70 \\
mono\_tok\_por & 39.5 & 40.3 & 31.9 & 53.5 & -4.20 & 48.3 & 41.2 & 32.1 & 56.8 & -10.8 & 11.68 & 9.11 & 4.02 & 15.18 & -10.0 & 3.40 & 3.00 & 0.70 & 6.90 & +0.00 \\
mono\_tok\_eng & 58.9 & 32.0 & 31.9 & 31.8 & -25.7 & 66.8 & 32.7 & 33.2 & 30.9 & -34.4 & 27.20 & 9.75 & 6.42 & 4.36 & -15.5 & 6.50 & 0.70 & 0.60 & 0.70 & -6.20 \\
\bottomrule
\end{tabular}
}
\end{table}

\begin{table}[h]
\centering
\caption{3B model results across benchmarks. Models prefixed with \texttt{mono\_tok} use monolingual tokenizers trained from scratch on each language. $\Delta$ measures the difference between the best non-English score for each model and the English-only baseline score at the same scale.}
\label{tab:mono_tok_3B}
\resizebox{\textwidth}{!}{
\begin{tabular}{lcccccccccccccccccccc}
\toprule
 & \multicolumn{5}{c}{HellaSwag} & \multicolumn{5}{c}{ARC-Easy} & \multicolumn{5}{c}{TQA} & \multicolumn{5}{c}{NQ} \\
\cmidrule(lr){2-6} \cmidrule(lr){7-11} \cmidrule(lr){12-16} \cmidrule(lr){17-21}
 & en & es & fr & pt & $\Delta$ & en & es & fr & pt & $\Delta$ & en & es & fr & pt & $\Delta$ & en & es & fr & pt & $\Delta$ \\
\midrule
eng\_only\_3B & 67.4 & 38.4 & 37.0 & 37.1 & +0.00 & 72.9 & 39.3 & 38.4 & 35.5 & +0.00 & 38.77 & 20.54 & 14.97 & 12.42 & +0.00 & 10.4 & 2.30 & 2.80 & 3.00 & +0.00 \\
mono\_tok\_spa\_3B & 49.3 & 63.0 & 36.7 & 45.1 & -4.40 & 55.6 & 67.0 & 37.2 & 46.0 & -5.90 & 19.56 & 24.88 & 11.93 & 15.25 & -13.9 & 5.30 & 7.50 & 3.00 & 5.40 & -2.90 \\
mono\_tok\_por\_3B & 47.4 & 45.2 & 35.4 & 60.5 & -6.90 & 52.2 & 47.8 & 36.7 & 59.2 & -13.7 & 10.89 & 12.67 & 4.54 & 12.72 & -26.1 & 3.70 & 2.50 & 1.30 & 8.10 & -2.30 \\
mono\_tok\_fra\_3B & 47.2 & 38.8 & 57.3 & 35.2 & -10.1 & 50.3 & 35.4 & 55.8 & 33.4 & -17.1 & 20.18 & 14.53 & 20.70 & 10.40 & -18.1 & 3.30 & 2.40 & 5.50 & 2.20 & -4.90 \\
mono\_tok\_eng\_3B & 67.6 & 38.0 & 36.9 & 36.4 & -29.4 & 72.3 & 40.0 & 39.1 & 35.7 & -32.9 & 40.49 & 19.51 & 15.63 & 12.03 & -19.3 & 9.70 & 2.40 & 3.50 & 2.20 & -6.90 \\
\bottomrule
\end{tabular}
}
\end{table}

\subsection{Controlled Experiment: Same Data, Fair Tokenizers} \label{appendix:fair_tokenizers_same_data}
By training both models on semantically equivalent corpora using tokenizers with matched compression ratios, we successfully isolate the models' acquired knowledge from structural training disparities. As shown in \Cref{tab:controlled}, this dual-control approach drastically reduces the massive performance disparities observed in the uncontrolled baselines. For instance, the Spanish ARC-Easy deficit drops to roughly 4--6\% across all scales, down from over 11\% in the single-control setups.
We focus our exploration to Spanish only due to the resources required to translate DCLM to other languages, as reported in \Cref{section:appendix/translation_of_training_data}.

However, the critical takeaway from this table is that a gap persists. Despite the models possessing identical underlying knowledge, the English evaluation consistently outscores the Spanish evaluation on HellaSwag (by 2.3\% to 4.0\%) and TriviaQA (by 6.1\% to 8.9\%). Interestingly, Natural Questions at the 8B scale presents a notable exception, where the Spanish model actually edges out the English model by +0.6\%. 

Because we have strictly matched both the knowledge seen during training and the efficiency of the tokenization, we question whether these gaps can be attributed to a genuine deficit in model capability. They serve as the foundational evidence for our main paper's thesis: the remaining disparities are measurement artifacts, driven by how the evaluation benchmarks query this knowledge via arbitrary surface forms.
\begin{table}[h]
    \centering
    \caption{Same data (LLM-translated DCLM-EDU), language-specific tokenizers with matched compression. Each model evaluated in its own language. $\Delta$ are computed with the Spanish model against the English model performance in English. Gaps persist across scales despite identical training data.}
    \label{tab:controlled}
\resizebox{\textwidth}{!}{
\begin{tabular}{llcccccccccccc}
\toprule
 & & \multicolumn{3}{c}{HellaSwag} & \multicolumn{3}{c}{ARC-Easy} & \multicolumn{3}{c}{TQA} & \multicolumn{3}{c}{NQ} \\
\cmidrule(lr){3-5} \cmidrule(lr){6-8} \cmidrule(lr){9-11} \cmidrule(lr){12-14}
Scale & & en & es & $\Delta$ & en & es & $\Delta$ & en & es & $\Delta$ & en & es & $\Delta$ \\
\midrule
\multirow{2}{*}{1B} & eng & 58.6 & 32.4 & --- & 66.6 & 32.1 & --- & 28.3 & 9.6 & --- & 7.2 & 1.1 & --- \\
& spa & 37.3 & 56.3 & --2.3 & 46.5 & 60.7 & --5.9 & 1.3 & 21.3 & --7.0 & 2.4 & 5.3 & --1.9 \\
\midrule
\multirow{2}{*}{3B} & eng & 68.1 & 36.6 & --- & 74.0 & 38.2 & --- & 40.6 & 21.3 & --- & 10.9 & 2.4 & --- \\
& spa & 42.1 & 64.1 & --4.0 & 52.2 & 69.9 & --4.1 & 3.3 & 31.7 & --8.9 & 3.8 & 9.7 & --1.2 \\
\midrule
\multirow{2}{*}{8B} & eng & 75.0 & 44.5 & --- & 78.5 & 42.8 & --- & 53.5 & 32.1 & --- & 15.2 & 5.2 & --- \\
& spa & 49.1 & 71.1 & --3.9 & 56.6 & 74.7 & --3.8 & 5.7 & 47.4 & --6.1 & 11.3 & 15.8 & +0.6 \\
\bottomrule
\end{tabular}
}
\end{table}

\section{\methodname{} implementation} \label{section:appendix/code}
in \Cref{alg:paraeval} we show the implementation of \methodname{} in lm-harness, which requires a simple change in the choices, by flattening all paraphrases into one list, and the targets, by allowing all paraphrases of the correct choice to be correct.
\begin{algorithm}[t]
\small
\caption{ParaEval via \texttt{lm-eval-harness}}
\label{alg:paraeval}
\begin{algorithmic}[1]
\Require MCQA dataset $\mathcal{D}=\{q_i\}_{i=1}^N$; each $q_i$ has $C$ choices; each choice has $K$ variants $v_{i,c,k}$
\Ensure ParaEval accuracy

\Statex \textbf{YAML configuration:}
\Statex \hspace{\algorithmicindent} \texttt{output\_type: multiple\_choice}
\Statex \hspace{\algorithmicindent} \texttt{doc\_to\_choice: flatten\_choices}
\Statex \hspace{\algorithmicindent} \texttt{doc\_to\_target: correct\_indices}
\Statex

\Function{flatten\_choices}{$q_i$}
    \State $flat \gets [\,]$
    \For{$c = 1$ \textbf{to} $C$}
        \For{$k = 1$ \textbf{to} $K$}
            \State \Call{Append}{$flat, v_{i,c,k}$}
        \EndFor
    \EndFor
    \State \Return $flat$
\EndFunction
\Statex

\Function{correct\_indices}{$q_i$}
    \State $gold \gets [\,]$
    \State $idx \gets 0$
    \For{$c = 1$ \textbf{to} $C$}
        \For{$k = 1$ \textbf{to} $K$}
            \If{$c = \text{correct}(q_i)$}
                \State \Call{Append}{$gold, idx$}
            \EndIf
            \State $idx \gets idx + 1$
        \EndFor
    \EndFor
    \State \Return $gold$
\EndFunction
\Statex

\Function{Score}{model, $q_i$, $flat$, $gold$}
    \For{$j = 1$ \textbf{to} $|flat|$}
        \State $ll_j \gets \log P_{\text{model}}(flat[j] \mid \text{prompt}(q_i))$
    \EndFor
    \State $pred \gets \arg\max_j ll_j$
    \State \Return $\mathbf{1}[pred \in gold]$
\EndFunction
\Statex

\State $acc \gets 0$
\For{$i = 1$ \textbf{to} $N$}
    \State $flat_i \gets \Call{flatten\_choices}{q_i}$
    \State $gold_i \gets \Call{correct\_indices}{q_i}$
    \State $acc \gets acc + \Call{Score}{\text{model}, q_i, flat_i, gold_i}$
\EndFor
\State \Return $acc / N$
\end{algorithmic}
\end{algorithm}

\section{Prompting}\label{app:prompting}
\begin{figure}[h]
    \centering
    \includegraphics[width=\columnwidth]{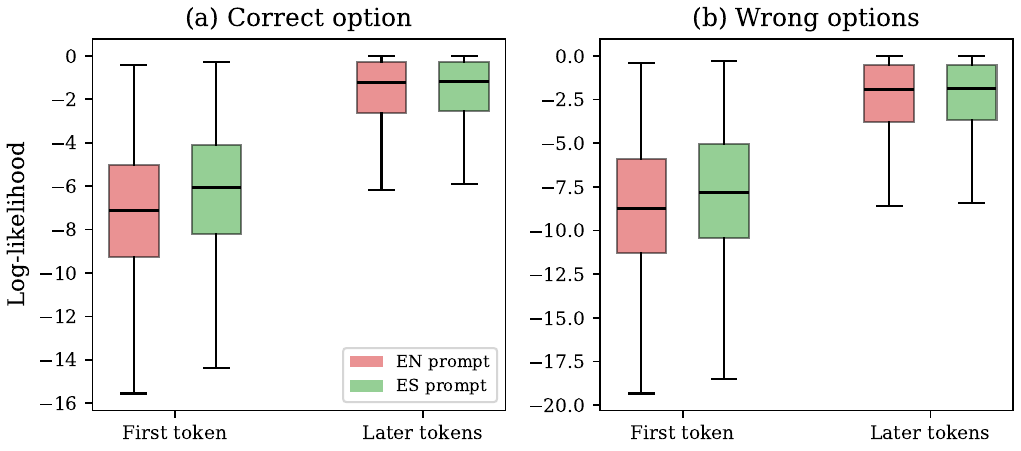}
    \caption{Per-token log-likelihood distributions for the Spanish model on ARC-Easy (ES), comparing English prompt (``Question/Answer'') vs.\ Spanish prompt (``Pregunta/Respuesta'') in a 0-shot setting. At the first completion token, the English prompt yields substantially lower log-likelihoods (median $\Delta = +1.03$ for correct options, $+0.89$ for wrong options), indicating higher perplexity. Later tokens are unaffected ($\Delta = +0.04$), showing that the prompt-language mismatch penalizes mostly the initial transition from prompt to completion.}
    \label{fig:per-token-perplexity}
\end{figure}

Standard evaluation harnesses use English prompt templates (e.g., ``Question: \{Q\} Answer:'') even for non-English tasks, e.g. Global-MMLU.
We find that switching the prompt to the evaluation language (e.g., ``Pregunta: \{Q\} Respuesta:'') drastically reduces the perplexity of the first tokens for the Spanish model, and improves the performance overall beacause of that.
In \Cref{fig:per-token-perplexity} we show how the language of the prompt affects the perplexity of the first tokens, skewing the evaluation of models.

We additionally noticed that the model, in MCQA under 0-shot native prompt, was failing several questions it had scored correctly in the generative version, see \Cref{tab:scoring_methods}, which makes sense given that 1B models have a hard time understanding what is the task at hand, and are not instruct-tuned with prompt following capabilities, so 5-shot gives them better context and calibration.
We verify the performance gains with native prompts, and with 5-shot in \Cref{tab:arce-breakdown}, where we can see the language of the prompt already reduces the gap dramatically.

\begin{table}[h]
\centering
\caption{ARC-Easy evaluation breakdown (1B). Progressively fairer evaluation reduces the EN--ES gap.}
\label{tab:arce-breakdown}
\begin{tabular}{l ccc}
\toprule
\textbf{Configuration} & \textbf{EN} & \textbf{ES} & \textbf{Gap} \\
\midrule
0-shot, English prompt (acc) & 71.6 & 66.3 & +5.3 \\
0-shot, English prompt (acc\_norm) & 66.6 & 60.7 & +5.9 \\
0-shot, Spanish prompt (acc) & 71.6 & 67.6 & +4.0 \\
0-shot, Spanish prompt (acc\_norm) & 66.6 & 62.8 & +3.9 \\
5-shot, native prompt (acc) & 72.3 & 70.6 & +1.8 \\
5-shot, native prompt (acc\_norm) & 72.4 & 68.2 & +4.3 \\
\quad + ParaEval (acc) & 70.4 & 70.7 & -0.3 \\
\quad + ParaEval (acc\_norm) & 67.9 & 65.4 & +2.5 \\
\bottomrule
\end{tabular}
\end{table}

\section{Translationese} \label{appendix:translationese}
We backtranslate benchmarks (EN$\to$ES$\to$EN) and evaluate the English model on both original and backtranslated versions (\Cref{tab:translationese}).
The English model itself loses 1.5-3.7\% on HellaSwag and 1.4-3.3\% on TQA from translationese alone, demonstrating that the different phrasing has a direct impact in performance.

\begin{table}[h]
\centering
\caption{Effect of translationese: English models evaluated on original vs.\ backtranslated (EN$\to$ES$\to$EN) benchmarks. Performance is measured with acc\_norm.}
\label{tab:translationese}
\resizebox{\textwidth}{!}{
\begin{tabular}{l ccc ccc ccc ccc}
\toprule
 & \multicolumn{3}{c}{HellaSwag} & \multicolumn{3}{c}{ARC-Easy} & \multicolumn{3}{c}{TQA} & \multicolumn{3}{c}{NQ} \\
\cmidrule(lr){2-4} \cmidrule(lr){5-7} \cmidrule(lr){8-10} \cmidrule(lr){11-13}
Model & en & en\_bt & $\Delta$ & en & en\_bt & $\Delta$ & en & en\_bt & $\Delta$ & en & en\_bt & $\Delta$ \\
\midrule
Eng-Only 1B & 57.7 & 56.2 & --1.5 & 67.6 & 67.0 & --0.5 & 25.2 & 24.1 & --1.1 & 6.9 & 7.4 & +0.5 \\
Eng-Only 3B & 67.4 & 65.0 & --2.4 & 72.9 & 72.5 & --0.4 & 38.8 & 36.8 & --2.0 & 10.4 & 11.7 & +1.3 \\
Eng-Only 8B & 74.7 & 71.0 & --3.7 & 78.4 & 77.1 & --1.3 & 51.8 & 48.7 & --3.1 & 14.8 & 15.7 & +0.9 \\
\bottomrule
\end{tabular}}
\end{table}

\section{Completion Paraphrase Metrics}\label{appendix:paravg}
\Cref{tab:mean_vs_max} investigates the effect of different aggregation strategies for scoring paraphrase variants.
Given $k$ paraphrases per answer choice, we compare three methods for aggregating their log-likelihoods into a single score: averaging ($\frac{1}{k}\sum_v \text{LL}_v$), taking the maximum ($\max_v \text{LL}_v$: ParaEval), and log-sum-exp ($\log\sum_v \exp(\text{LL}_v)$).
Results demonstrate that while averaging consistently fails to close the cross-lingual performance gap, both the maximum and log-sum-exp aggregations successfully mitigate it.
Between the two, the maximum yields the smallest EN--ES gaps overall, particularly on ARC where log-sum-exp still leaves residual gaps of up to +1.37pp.

\begin{table}[h]
\centering
\caption{ParaEval (max) vs.\ Mean LL vs.\ LogSumExp aggregation on ARC and HellaSwag (5-shot, raw LL). Mean LL averages log-likelihoods across paraphrase variants per choice, ParaEval selects the maximum, and LogSumExp computes $\log\sum\exp(\text{LL})$. Results at 1B include $\pm$std over 3 seeds.}
\label{tab:mean_vs_max}
\begin{tabular}{ll ccc}
\toprule
\textbf{Benchmark / Scale} & \textbf{Method} & \textbf{EN} & \textbf{ES} & \textbf{Gap} \\
\midrule
\multicolumn{5}{l}{\textit{ARC}} \\
\midrule
\multirow{3}{*}{1B} & ParaEval (raw LL) & 70.52$\pm$0.68 & 70.24$\pm$0.33 & +0.28 \\
                     & Mean LL (raw)     & 67.23$\pm$0.35 & 62.39$\pm$0.65 & +4.84 \\
                     & LogSumExp     & 71.17$\pm$0.93 & 70.59$\pm$0.40 & +0.58 \\
\cmidrule(lr){1-5}
\multirow{3}{*}{3B} & ParaEval (raw LL) & 48.29 & 47.78 & +0.51 \\
                     & Mean LL (raw)     & 45.39 & 39.42 & +5.97 \\
                     & LogSumExp     & 48.89 & 47.53 & +1.37 \\
\cmidrule(lr){1-5}
\multirow{3}{*}{8B} & ParaEval (raw LL) & 56.70 & 56.20 & +0.50 \\
                     & Mean LL (raw)     & 52.30 & 48.21 & +4.10 \\
                     & LogSumExp     & 57.34 & 56.23 & +1.11 \\
\midrule
\multicolumn{5}{l}{\textit{HellaSwag}} \\
\midrule
\multirow{3}{*}{1B} & ParaEval (raw LL) & 41.33$\pm$0.15 & 41.49$\pm$0.22 & $-$0.16 \\
                     & Mean LL (raw)     & 39.20$\pm$0.16 & 38.98$\pm$0.13 & +0.22 \\
                     & LogSumExp     & 41.29$\pm$0.14 & 41.47$\pm$0.21 & $-$0.18 \\
\cmidrule(lr){1-5}
\multirow{3}{*}{3B} & ParaEval (raw LL) & 46.40 & 46.10 & +0.30 \\
                     & Mean LL (raw)     & 43.55 & 42.75 & +0.80 \\
                     & LogSumExp     & 46.57 & 46.17 & +0.40 \\
\cmidrule(lr){1-5}
\multirow{3}{*}{8B} & ParaEval (raw LL) & 51.60 & 51.10 & +0.50 \\
                     & Mean LL (raw)     & 48.03 & 46.79 & +1.24 \\
                     & LogSumExp     & 51.53 & 51.13 & +0.40 \\
\bottomrule
\end{tabular}
\end{table}

\section{Evaluation costs}\label{appendix:evaluation_costs}
\Cref{tab:benchmark_expansion} provides a detailed breakdown of the token expansion relative to the original benchmark sizes.
\methodname{} adds inference cost ($k{+}1$ forward passes per option) but requires no retraining and applies to any MCQA benchmark, multilingual or monolingual. Across the three evaluated benchmarks, the method yields a corpus approximately three times larger in total token count. Assuming a fully optimized inference pipeline utilizing KV-caching for both the prompt and few-shot exemplars, this token expansion theoretically bounds the execution time increase to a factor of $3\times$. However, in the absence of such caching mechanisms, the computational bottleneck remains dominated by the initial processing of the prompt and few-shot context. Empirically, when evaluating within the \texttt{lm-evaluation-harness} (which does not cache the fewshot or the prompt), we observe the mean wall-clock times shifting from 25s to 36s for 1B parameter models, 21s to 41s for 3B models, and 23s to 55s for 8B models, ran 3 times over 8 H100 gpus.
\begin{table}[htbp]
\centering
\caption{Benchmark Token Counts and Expansion Metrics}
\label{tab:benchmark_expansion}
\resizebox{\textwidth}{!}{%
\begin{tabular}{llrrrrrrrr}
\toprule
\textbf{Benchmark} & \textbf{Lang} & \textbf{Fewshot} & \textbf{Prompt} & \textbf{Orig Compl} & \textbf{Div Compl} & \textbf{Expansion} & \textbf{Corpus} & \textbf{Compl Size} & \textbf{Total Data} \\
& & \textbf{Toks} & \textbf{Tokens} & \textbf{Tokens} & \textbf{Tokens} & & \textbf{Tokens} & \textbf{Exp (\%)} & \textbf{Exp (\%)} \\
\midrule
ARC-E     & EN & 107 & 61,758  & 42,224    & 251,877   & 1.4013 & 103,982   & 596.53\% & 301.62\% \\
ARC-E     & ES & 133 & 72,733  & 48,317    & 291,768   & 1.4186 & 121,050   & 603.86\% & 301.12\% \\
ARC-C     & EN & 138 & 34,695  & 27,478    & 160,637   & 1.1692 & 62,173    & 584.60\% & 314.17\% \\
ARC-C     & ES & 175 & 40,235  & 30,801    & 176,932   & 1.1489 & 71,036    & 574.44\% & 305.71\% \\
Hellaswag & EN & 188 & 480,200 & 1,105,108 & 5,353,855 & 0.9689 & 1,623,715 & 484.46\% & 359.30\% \\
Hellaswag & ES & 196 & 494,380 & 1,119,636 & 5,588,802 & 0.9983 & 1,662,441 & 499.16\% & 365.92\% \\
\bottomrule
\end{tabular}%
}
\end{table}

\section{Analysis of prediction changes} \label{section:appendix/paraeval_analysis_of_prediction_changes}
In this section we explore the loglikelihood changes introduced by \methodname{}, to better understand what is happening, and what is driving the performance differences.
\subsection{Qualitative}
To understand how ParaEval changes predictions we examine ARC-Easy (1B) questions where the answer flips between Original and ParaEval. In English, 102 flip wrong$\to$right and 149 flip right$\to$wrong (net $-1.9$). In Spanish the counts are balanced, 126 vs.\ 124 (net $+0.1$). Both directions reflect the same phenomenon: sensitivity to surface form, not a uniform accuracy gain. The net change therefore depends on how well the original phrasing matched the model's training distribution, not on the language.

Because we score with raw summed log-likelihood and take the maximum over paraphrases, a natural concern is that ParaEval could win simply by introducing a shorter variant for each option, since shorter strings accumulate fewer negative log-probs. Three checks show this is not the main driver. First, ParaEval selects the shortest choice in 33.8\% of English questions and 32.1\% in Spanish, essentially identical to Original (33.8\% and 33.4\%). Second, an always-shortest policy scores $\sim$25\%, near chance and far below both methods ($>$70\%). Third, the pattern persists under length-normalized scoring (\texttt{acc\_norm}): the English gap remains negative ($-1.3$) and the Spanish gap positive ($+0.9$). Moreover, 25.5\% of English right$\to$wrong flips (37.1\% in Spanish) occur when the winning distractor is the same length or longer than the correct answer's best variant.

Table~\ref{tab:paraeval-flips} shows representative examples from both languages and both flip directions. Examining them reveals consistent patterns in how surface form alters model preference.

\textbf{Wrong$\to$Right} flips recover knowledge hidden by unnatural wording. In English \textit{ecological relationships}, ``parasite-host'' ($-4.2$) loses to the more natural ``parasitism'' ($-0.5$), flipping the prediction away from ``mutualism.'' In \textit{mineral hardness}, ``Talc is softer than diamonds'' ($-18.5$) is rescued by the rephrasing ``Diamond is harder than talc'' ($-10.5$). In Spanish the effect is clearer because the paraphrases collapse to single words: in \textit{relaciones ecológicas} all options become 11-character adjectives, yet the model clearly prefers ``Cooperativa'' ($-9.2$) over ``Parasitaria'' ($-10.8$); in \textit{herencia genética}, ``transmitidos por dos progenitores'' ($-19.2$) is recovered by ``De ambos progenitores'' ($-11.7$).

\textbf{Right$\to$Wrong} flips expose original wins that relied on a phrasing advantage. In \textit{speed at depth}, the correct ``They slow down'' and the distractor ``They speed up'' are rewritten to the parallel forms ``Their speed decreases'' ($-8.2$) and ``Their speed increases'' ($-8.1$); the model flips on virtually identical surface forms, indicating it does not robustly distinguish the outcomes. In \textit{genetics}, the distractor ``Genes are able to disappear and reappear in later generations'' ($-30.3$) becomes the concise and familiar ``Genes can skip generations'' ($-12.7$), overtaking the correct ``Genes control inherited traits in organisms.'' Spanish shows the same two modes: in \textit{propiedades de ondas}, ``Su altura aumenta'' and ``Su altura disminuye'' map to the symmetric ``Aumenta su altura'' ($-9.3$) and ``Disminuye su altura'' ($-8.0$); in \textit{dirección vs.\ distancia}, the verbose distractor ``la cantidad de distancia recorrida durante el movimiento'' ($-20.2$) collapses to ``distancia recorrida'' ($-3.5$) and beats the correct ``la dirección del movimiento.''

This explains the asymmetric net effects reported in the main results. English scores fall slightly because several correct answers originally benefited from concise, high-frequency phrasing (e.g., ``mutualism'' beating ``parasite-host''), while Spanish scores remain neutral because the translated options were on average more verbose and less idiomatic, so \methodname{} more often recovers suppressed knowledge (e.g., ``Cooperativa'' vs.\ ``La relación es cooperativa''). ParaEval thus acts as a denoiser: it does not systematically improve accuracy, it reveals where accuracy was previously inflated or deflated by surface form.

The same dynamics appear on harder tasks and larger models. Table~\ref{tab:paraeval-flips-arcc} shows representative flips for ARC-Challenge (3B), where ParaEval improves both English and Spanish. The mechanisms are identical: in \textit{electric charge}, the model recovers the correct ``charge of -3'' once paraphrased as ``negative three charge'' ($-6.5$ vs.\ $-9.3$); in \textit{sound and electricity}, the original correct ``converting sound to electrical impulses'' loses to the symmetric distractor ``converting electrical signals into sound'' ($-10.5$ vs.\ $-12.1$). Spanish shows the same pattern, with \textit{deposición fluvial} flipping to the more natural ``acumulación de sedimentos en el cauce,'' and \textit{tipos de onda} flipping from ``longitudinal'' to the longer but more frequent ``onda transversal.'' The fact that both languages gain here, while English lost on ARC-Easy (1B), underscores that ParaEval does not have a built-in direction: it simply removes the phrasing advantage whichever side originally held it.

We additionally check the EN-BT model, examples under \Cref{tab:paraeval-flips-bt}, and check that it also has the tendency to prefer the canonical name instead of an explanation as ``The mountain was submerged"" is selected over the original ``The mountain was covered by water"". 

\begin{table}[h!]
\centering
\caption{Representative prediction flips between Original and \methodname{} on ARC-Easy (1B). \textbf{W$\to$R:} the model possesses the knowledge but the original phrasing suppresses it. \textbf{R$\to$W:} the original prediction relied on a surface-form advantage rather than genuine knowledge. \checkmark\,= correct answer; $\leftarrow$\,= selected answer.}
\label{tab:paraeval-flips}
\resizebox{\textwidth}{!}{
\begin{tabular}{cl rr rr l}
\toprule
& \textbf{Choice (original text)} & \multicolumn{2}{c}{\textbf{Original}} & \multicolumn{2}{c}{\textbf{\methodname{}}} & \textbf{Best variant} \\
\cmidrule(lr){3-4} \cmidrule(lr){5-6}
& & LL & Sel & LL & Sel & \\
\midrule
\multicolumn{7}{c}{\textbf{English}} \\
\midrule
\multicolumn{7}{l}{\textit{Ecological relationships} (W$\to$R)} \\
& predator-prey & -5.2 &  & -5.2 &  & --- \\
& mutualism & -3.6 & $\leftarrow$ & -3.6 &  & --- \\
& \checkmark parasite-host & -4.2 &  & -0.5 & $\leftarrow$ & parasitism \\
& commensalism & -4.2 &  & -4.2 &  & --- \\
\cmidrule(lr){1-7}
\multicolumn{7}{l}{\textit{Mineral hardness} (W$\to$R)} \\
& Diamonds have sharp edges & -16.2 & $\leftarrow$ & -16.2 &  & --- \\
& \checkmark Talc is softer than diamonds & -18.5 &  & -10.5 & $\leftarrow$ & Diamond is harder than talc \\
& Diamonds are very valuable & -23.9 &  & -23.4 &  & A diamond is worth a lot of money \\
& Talc and diamonds are both minerals & -21.0 &  & -20.5 &  & Both talc and diamonds are minerals \\
\cmidrule(lr){1-7}
\multicolumn{7}{l}{\textit{Mineral identification} (W$\to$R)} \\
& Weigh the mineral & -11.6 & $\leftarrow$ & -11.6 &  & --- \\
& Dip the mineral in acid & -13.9 &  & -13.7 &  & Acid dip \\
& Heat the mineral until it melts & -14.7 &  & -12.1 &  & Melt it \\
& \checkmark Drag an edge of the mineral across a tile & -28.9 &  & -10.6 & $\leftarrow$ & Streak test \\
\cmidrule(lr){1-7}
\multicolumn{7}{l}{\textit{Speed at depth} (R$\to$W)} \\
& \checkmark They slow down & -9.9 & $\leftarrow$ & -8.2 &  & Their speed decreases \\
& They speed up & -12.5 &  & -8.1 & $\leftarrow$ & Their speed increases \\
& They decrease in number & -13.6 &  & -13.6 &  & --- \\
& They decrease in size & -11.6 &  & -9.2 &  & They get smaller \\
\cmidrule(lr){1-7}
\multicolumn{7}{l}{\textit{Genetics} (R$\to$W)} \\
& \checkmark Genes control inherited traits in organisms & -20.1 & $\leftarrow$ & -17.4 &  & Genes govern heritable traits \\
& Corn plants can reproduce in laboratory environments & -34.5 &  & -29.5 &  & Corn plants can be made to reproduce under l\ldots \\
& Chromosomes can move around in the nucleus of a cell & -20.5 &  & -14.9 &  & Chromosomes move within the cell nucleus \\
& Genes are able to disappear and reappear in later g\ldots & -30.3 &  & -12.7 & $\leftarrow$ & Genes can skip generations \\
\cmidrule(lr){1-7}
\multicolumn{7}{l}{\textit{Direction vs.\ distance} (R$\to$W)} \\
& the rate of the motion & -12.1 &  & -8.7 &  & speed of movement \\
& \checkmark the direction of the motion & -7.2 & $\leftarrow$ & -7.2 &  & --- \\
& the change in the amount of motion & -18.0 &  & -16.7 &  & how much the motion changes \\
& the amount of distance traveled during motion & -20.8 &  & -4.4 & $\leftarrow$ & distance traveled \\
\midrule
\multicolumn{7}{c}{\textbf{Spanish}} \\
\midrule
\multicolumn{7}{l}{\textit{Relaciones ecol\'{o}gicas} (W$\to$R)} \\
& La relaci\'{o}n es competitiva & -21.1 &  & -13.9 &  & Competitiva \\
& \checkmark La relaci\'{o}n es cooperativa & -22.2 &  & -9.2 & $\leftarrow$ & Cooperativa \\
& La relaci\'{o}n es parasitaria & -16.5 & $\leftarrow$ & -10.8 &  & Parasitaria \\
& La relaci\'{o}n es depredadora & -19.6 &  & -13.1 &  & Depredadora \\
\cmidrule(lr){1-7}
\multicolumn{7}{l}{\textit{Herencia gen\'{e}tica} (W$\to$R)} \\
& heredados de un solo progenitor & -14.3 & $\leftarrow$ & -13.3 &  & De un solo progenitor \\
& creados por el medio ambiente & -21.7 &  & -16.3 &  & Producto del ambiente \\
& \checkmark transmitidos por dos progenitores & -19.2 &  & -11.7 & $\leftarrow$ & De ambos progenitores \\
& aprendidos de los hermanos & -28.1 &  & -17.6 &  & De los hermanos \\
\cmidrule(lr){1-7}
\multicolumn{7}{l}{\textit{Medici\'{o}n astron\'{o}mica} (W$\to$R)} \\
& \checkmark midendo su corrimiento al rojo & -31.9 &  & -9.7 & $\leftarrow$ & Corrimiento al rojo \\
& la gravedad que ejerce & -18.6 &  & -15.9 &  & Su gravedad \\
& compar\'{a}ndola con estrellas cercanas & -18.3 &  & -18.3 &  & --- \\
& observando su movimiento a trav\'{e}s del cielo & -14.6 & $\leftarrow$ & -14.6 &  & --- \\
\cmidrule(lr){1-7}
\multicolumn{7}{l}{\textit{Propiedades de ondas} (R$\to$W)} \\
& \checkmark Su altura aumenta & -15.0 & $\leftarrow$ & -9.3 &  & Aumenta su altura \\
& Su altura disminuye & -15.1 &  & -8.0 & $\leftarrow$ & Disminuye su altura \\
& Su longitud se acorta & -19.2 &  & -11.2 &  & La longitud de onda se vuelve m\'{a}s corta \\
& Se forma la cresta & -22.7 &  & -12.9 &  & La cresta se forma \\
\cmidrule(lr){1-7}
\multicolumn{7}{l}{\textit{Direcci\'{o}n vs.\ distancia} (R$\to$W)} \\
& la rapidez del movimiento & -9.8 &  & -9.8 &  & --- \\
& \checkmark la direcci\'{o}n del movimiento & -7.6 & $\leftarrow$ & -7.6 &  & --- \\
& el cambio en la cantidad de movimiento & -15.9 &  & -14.4 &  & variaci\'{o}n del momento \\
& la cantidad de distancia recorrida durante el movim\ldots & -20.2 &  & -3.5 & $\leftarrow$ & distancia recorrida \\
\cmidrule(lr){1-7}
\multicolumn{7}{l}{\textit{Biolog\'{i}a celular} (R$\to$W)} \\
& forman sistemas & -14.0 &  & -13.4 &  & dan lugar a sistemas \\
& \checkmark est\'{a}n hechos de c\'{e}lulas & -12.6 & $\leftarrow$ & -7.0 &  & c\'{e}lulas \\
& est\'{a}n hechos de \'{o}rganos & -19.3 &  & -6.4 & $\leftarrow$ & \'{o}rganos \\
& realizan el trabajo de las c\'{e}lulas & -17.4 &  & -17.1 &  & llevan a cabo las tareas de las c\'{e}lulas \\
\bottomrule
\end{tabular}}
\end{table}

\begin{table}[t]
\centering
\caption{Representative prediction flips between Original and \methodname{} on ARC-Challenge (3B). \textbf{W$\to$R:} the model possesses the knowledge but the original phrasing suppresses it. \textbf{R$\to$W:} the original prediction relied on a surface-form advantage rather than genuine knowledge. \checkmark\,= correct answer; $\leftarrow$\,= selected answer.}
\label{tab:paraeval-flips-arcc}
\resizebox{\textwidth}{!}{
\begin{tabular}{cl rr rr l}
\toprule
& \textbf{Choice (original text)} & \multicolumn{2}{c}{\textbf{Original}} & \multicolumn{2}{c}{\textbf{\methodname{}}} & \textbf{Best variant} \\
\cmidrule(lr){3-4} \cmidrule(lr){5-6}
& & LL & Sel & LL & Sel & \\
\midrule
\multicolumn{7}{c}{\textbf{English}} \\
\midrule
\multicolumn{7}{l}{\textit{Electric charge} (W$\to$R)} \\
& mass of +3 & -19.0 &  & -19.0 &  & --- \\
& mass of -3 & -17.9 &  & -17.9 &  & --- \\
& charge of +3 & -9.3 & $\leftarrow$ & -8.6 &  & charge of plus three \\
& \checkmark charge of -3 & -9.3 &  & -6.5 & $\leftarrow$ & negative three charge \\
\cmidrule(lr){1-7}
\multicolumn{7}{l}{\textit{Sound and electricity} (R$\to$W)} \\
& \checkmark converting sound to electrical impulses & -12.1 & $\leftarrow$ & -12.1 &  & --- \\
& receiving transmitted electrical signals & -18.5 &  & -18.5 &  & --- \\
& changing electrical impulses into sound & -13.5 &  & -10.5 & $\leftarrow$ & converting electrical signals into sound \\
& sending electrical signals through a circuit & -17.8 &  & -17.8 &  & --- \\
\midrule
\multicolumn{7}{c}{\textbf{Spanish}} \\
\midrule
\multicolumn{7}{l}{\textit{Deposición fluvial} (W$\to$R)} \\
& erosión de la orilla del arroyo & -17.4 & $\leftarrow$ & -17.4 &  & --- \\
& \checkmark deposición de material en la corriente & -24.0 &  & -17.1 & $\leftarrow$ & acumulación de sedimentos en el cauce \\
& cantidad de material transportado aguas abajo & -17.6 &  & -17.6 &  & --- \\
& tamaño de las partículas transportadas aguas abajo & -20.0 &  & -20.0 &  & --- \\
\cmidrule(lr){1-7}
\multicolumn{7}{l}{\textit{Tipos de onda} (R$\to$W)} \\
& electromagnético & -10.8 &  & -8.1 &  & una onda electromagnética \\
& calor & -10.3 &  & -10.3 &  & --- \\
& \checkmark longitudinal & -2.5 & $\leftarrow$ & -2.5 &  & --- \\
& transversal & -3.3 &  & -2.3 & $\leftarrow$ & onda transversal \\
\bottomrule
\end{tabular}}
\end{table}

\begin{table}[t]
\centering
\caption{Prediction flips for the standard English (Std) and backtranslated English (BT) models on ARC-Easy (1B). Both models converge to the correct answer under ParaEval once canonical phrasings are found, but the BT model requires paraphrasing more often because its training distribution penalizes natural English constructions. \checkmark\,= correct; $\leftarrow$\,= selected.}
\label{tab:paraeval-flips-bt}
\resizebox{\textwidth}{!}{
\begin{tabular}{cl rr rr rr rr}
\toprule
& & \multicolumn{4}{c}{\textbf{Standard model}} & \multicolumn{4}{c}{\textbf{BT model}} \\
\cmidrule(lr){3-6} \cmidrule(lr){7-10}
& \textbf{Choice (original text)} & \multicolumn{2}{c}{Original} & \multicolumn{2}{c}{ParaEval} & \multicolumn{2}{c}{Original} & \multicolumn{2}{c}{ParaEval} \\
\cmidrule(lr){3-4} \cmidrule(lr){5-6} \cmidrule(lr){7-8} \cmidrule(lr){9-10}
& & LL & Sel & LL & Sel & LL & Sel & LL & Sel \\
\midrule
\multicolumn{10}{l}{\textit{Disease prevention} --- Std: W$\to$R, BT: W$\to$R. Best variant: \emph{vaccination}} \\
& \checkmark getting a vaccination & $-$7.6 &  & $-$4.5 & $\leftarrow$ & $-$16.2 &  & $-$1.6 & $\leftarrow$ \\
& taking antibiotics & $-$9.1 &  & $-$8.5 &  & $-$9.3 & $\leftarrow$ & $-$7.9 &  \\
& eating fruits and vegetables & $-$10.5 &  & $-$10.5 &  & $-$12.3 &  & $-$12.3 &  \\
& washing hands often & $-$5.5 & $\leftarrow$ & $-$5.5 &  & $-$11.1 &  & $-$7.5 &  \\
\midrule
\multicolumn{10}{l}{\textit{Fossils on mountains} --- Std: stayed right, BT: W$\to$R. Best variant: \emph{The mountain was submerged}} \\
& The climate was once colder & $-$18.8 &  & $-$16.6 &  & $-$16.6 & $\leftarrow$ & $-$15.9 &  \\
& The shells were carried by the wind & $-$19.7 &  & $-$12.3 &  & $-$19.6 &  & $-$16.1 &  \\
& \checkmark The mountain was covered by water & $-$17.8 & $\leftarrow$ & $-$10.3 & $\leftarrow$ & $-$17.7 &  & $-$10.8 & $\leftarrow$ \\
& The shells once lived on dry land & $-$23.0 &  & $-$13.9 &  & $-$22.1 &  & $-$11.2 &  \\
\bottomrule
\end{tabular}}
\end{table}

\subsection{Quantitative}
We now quantify the effect of paraphrasing on the log-likelihood distributions of correct and wrong answer choices across three model scales and both languages.

\begin{figure}[h]
    \centering
    \includegraphics[width=0.6\columnwidth]{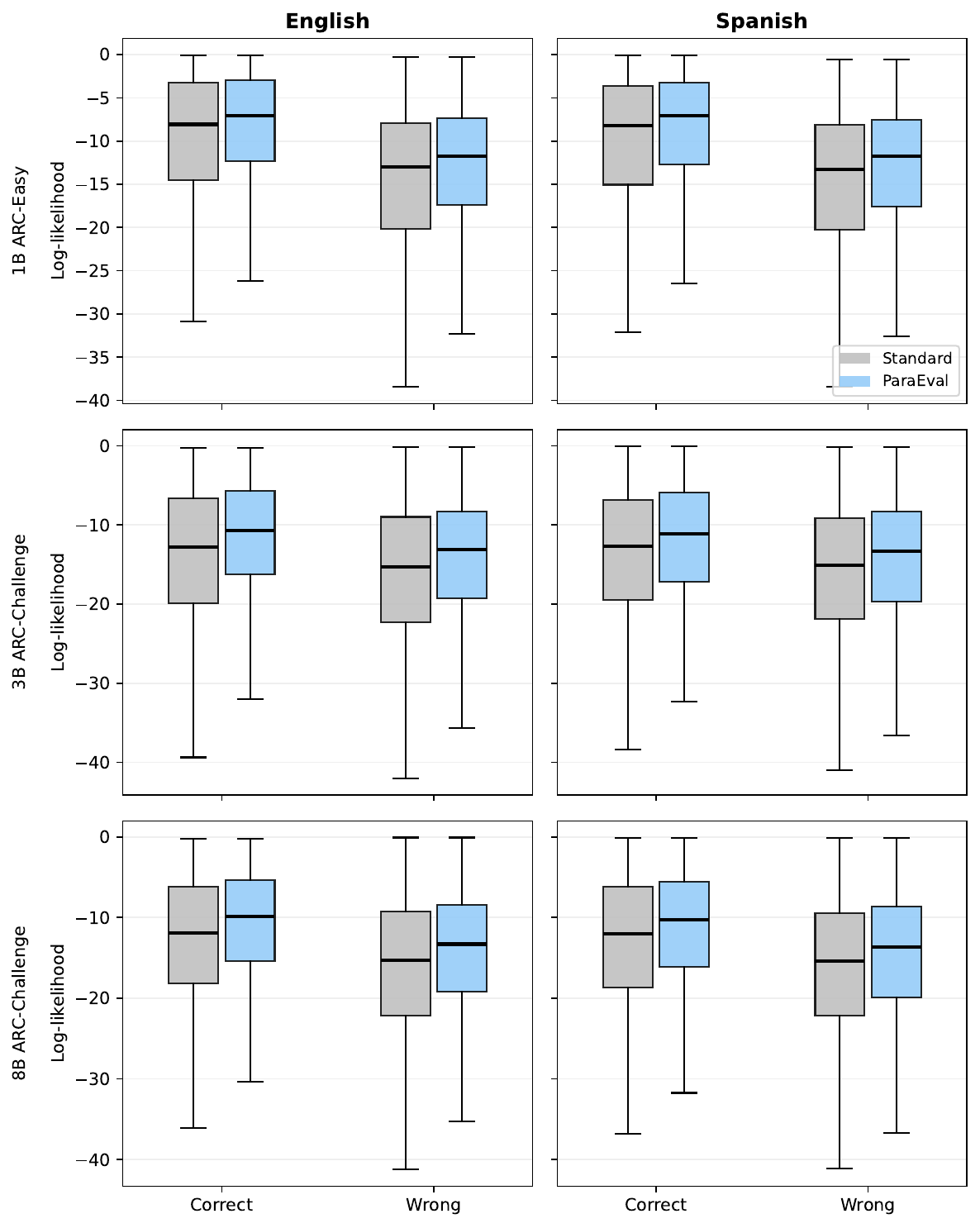}
    \caption{Distribution of raw log-likelihoods for correct and wrong answer choices under Standard (original phrasing only) and \methodname{} (best paraphrase) scoring. \methodname{} shifts both distributions upward by a similar amount, indicating that paraphrasing does not selectively advantage wrong options. Outliers omitted for clarity.}
    \label{fig:paraeval_ll_box_plot}
\end{figure}

\Cref{fig:paraeval_ll_box_plot} shows that \methodname{} shifts the log-likelihood distributions of both correct and wrong choices upward relative to Standard scoring. This confirms that paraphrasing does not systematically collapse the discriminability between correct and incorrect options, it raises all choices toward their most natural surface form.

\begin{figure}[h]
    \centering
    \includegraphics[width=0.8\columnwidth]{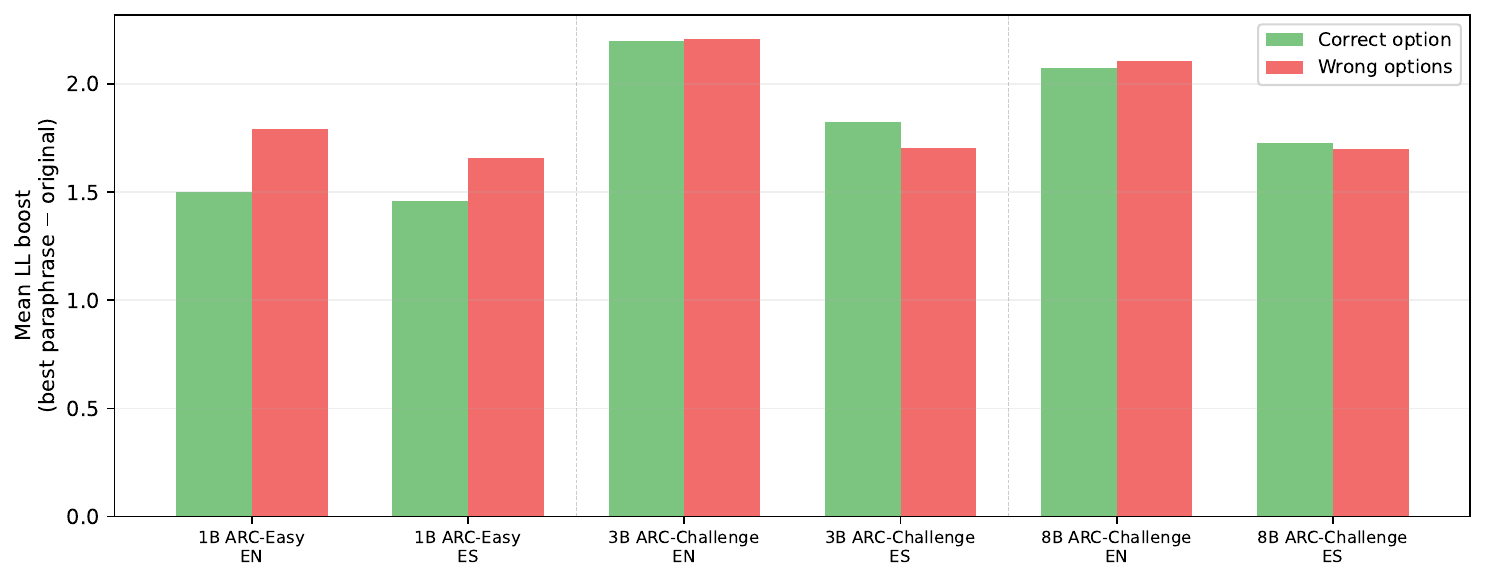}
    \caption{Mean log-likelihood boost from paraphrasing (best variant minus original) for correct and wrong answer choices. The boost is nearly symmetric across all settings, with slight asymmetries explaining the net accuracy changes: on 1B ARC-Easy EN, wrong options benefit marginally more ($+1.79$ vs.\ $+1.50$), consistent with the small accuracy drop under \methodname{}, while on ARC-Challenge ES, the pattern reverses, consistent with the accuracy gain.}
    \label{fig:paraeval_ll_diffs}
\end{figure}

\Cref{fig:paraeval_ll_diffs} makes this symmetry precise by comparing the mean boost (best paraphrase minus original log-likelihood) for correct and wrong options. Across all six settings the boost is nearly balanced, typically differing by less than 0.3 LL points. The direction of the small residual asymmetry directly predicts the net accuracy change: on 1B ARC-Easy EN, wrong options receive a slightly larger mean boost ($+1.79$) than correct ones ($+1.50$), consistent with the $-1.9$ accuracy drop under \methodname{}. On ARC-Challenge ES, correct options benefit marginally more ($+1.83$ vs.\ $+1.71$), consistent with the $+4.0$ accuracy gain. These small imbalances reflect differences in how naturally the original benchmark phrased correct versus wrong options, not a systematic bias introduced by \methodname{}.

\section{Paraphrasing of answers}

\subsection{Paraphrasing Model}\label{section:other_models_diversification}
In order to assess that our results are not limited to the generation style of Claude, we reprocess ARC-E, ARC-C, MMLU and HellaSwag with two additional models, Gemini 3.5 Pro and GPT-5. 
\Cref{tab:arce-diversified-all,tab:hellaswag-diversified-all,tab:bt-arce-diversified,tab:bt-hswag-diversified}
report results for ARC-Easy (1B), ARC-Challenge (3B/8B), HellaSwag (1B/3B/8B), and the EN vs.\ backtranslated-EN controls. All tables show the same pattern: ParaEval consistently reduces gaps across all three paraphrase generators, with absolute reductions of 1.5--2.5 on ARC and 0.3--1.2 on HellaSwag, confirming that the effect is not Claude-specific.

\begin{table}[t]
\centering
\caption{ARC diversified choices evaluation, Easy for 1B and Challenge for 3B and 8B.}
\label{tab:arce-diversified-all}
\resizebox{\textwidth}{!}{
\begin{tabular}{l ccc ccc ccc}
\toprule
& \multicolumn{3}{c}{\textbf{Claude}} & \multicolumn{3}{c}{\textbf{GPT}} & \multicolumn{3}{c}{\textbf{Gemini}} \\
\cmidrule(lr){2-4} \cmidrule(lr){5-7} \cmidrule(lr){8-10}
\textbf{Method} & EN & ES & Gap & EN & ES & Gap & EN & ES & Gap \\
\midrule
\multicolumn{10}{c}{1B} \\
Original (raw LL) & 72.3 & 70.6 & +1.8 & 72.3 & 70.6 & +1.8 & 72.3 & 70.6 & +1.8 \\
ParaEval (raw LL) & 70.4 & 70.7 & -0.3 & 70.4 & 71.1 & -0.7 & 71.1 & 70.6 & +0.5 \\
Paraphrasing correct (raw LL) & 79.9 & 78.8 & +1.1 & 81.0 & 79.0 & +2.0 & 79.7 & 76.9 & +2.8 \\
\midrule
\midrule
\multicolumn{10}{c}{3B} \\
Original (raw LL) & 46.9 & 45.1 & +1.9 & 46.9 & 45.1 & +1.9 & 46.9 & 45.1 & +1.9 \\
ParaEval (raw LL) & 48.3 & 47.8 & +0.5 & 49.6 & 49.1 & +0.5 & 48.9 & 48.7 & +0.2 \\
Paraphrasing correct (raw LL) & 61.3 & 56.9 & +4.4 & 62.4 & 58.1 & +4.3 & 59.4 & 51.9 & +7.5 \\
\midrule
\midrule
\multicolumn{10}{c}{8B} \\
Original (raw LL) & 54.4 & 53.6 & +0.8 & 54.4 & 53.6 & +0.8 & 54.4 & 53.6 & +0.8 \\
ParaEval (raw LL) & 56.7 & 56.2 & +0.5 & 58.7 & 58.4 & +0.3 & 56.9 & 56.7 & +0.2 \\
Paraphrasing correct (raw LL) & 68.2 & 65.2 & +3.0 & 69.5 & 64.1 & +5.4 & 65.4 & 59.5 & +5.9 \\
\bottomrule
\end{tabular}}
\end{table}

\begin{table}[t]
\centering
\caption{HellaSwag diversified choices evaluation.}
\label{tab:hellaswag-diversified-all}
\resizebox{\textwidth}{!}{
\begin{tabular}{l ccc ccc ccc}
\toprule
& \multicolumn{3}{c}{\textbf{Claude}} & \multicolumn{3}{c}{\textbf{GPT}} & \multicolumn{3}{c}{\textbf{Gemini}} \\
\cmidrule(lr){2-4} \cmidrule(lr){5-7} \cmidrule(lr){8-10}
\textbf{Method} & EN & ES & Gap & EN & ES & Gap & EN & ES & Gap \\
\midrule
\multicolumn{10}{c}{1B} \\
Original (raw LL) & 44.5 & 43.7 & +0.8 & 44.5 & 43.7 & +0.8 & 44.5 & 43.7 & +0.8 \\
ParaEval (raw LL) & 41.5 & 41.8 & -0.3 & 42.8 & 42.2 & +0.6 & 44.1 & 43.3 & +0.8 \\
Paraphrasing correct (raw LL) & 50.2 & 47.9 & +2.2 & 59.0 & 50.1 & +8.8 & 48.0 & 44.6 & +3.4 \\

\midrule
\midrule
\multicolumn{10}{c}{3B} \\
Original (raw LL) & 49.4 & 48.1 & +1.3 & 49.4 & 48.1 & +1.3 & 49.4 & 48.1 & +1.3 \\
ParaEval (raw LL) & 46.4 & 46.1 & +0.3 & 48.3 & 47.1 & +1.2 & 49.3 & 48.2 & +1.0 \\
Paraphrasing correct (raw LL) & 55.3 & 52.7 & +2.6 & 64.2 & 55.1 & +9.0 & 53.5 & 49.5 & +4.0 \\

\midrule
\midrule
\multicolumn{10}{c}{8B} \\
Original (raw LL) & 54.6 & 53.0 & +1.6 & 54.6 & 53.0 & +1.6 & 54.6 & 53.0 & +1.6 \\
ParaEval (raw LL) & 51.6 & 51.1 & +0.5 & 53.8 & 52.4 & +1.4 & 54.9 & 53.4 & +1.5 \\
Paraphrasing correct (raw LL) & 60.5 & 56.9 & +3.6 & 69.2 & 59.4 & +9.8 & 59.3 & 54.5 & +4.8 \\
\bottomrule
\end{tabular}}
\end{table}

\begin{table}[t]
\centering
\caption{MMLU diversified choices evaluation.}
\label{tab:mmlu-diversified-all}
\resizebox{\textwidth}{!}{
\begin{tabular}{l ccc ccc ccc}
\toprule
& \multicolumn{3}{c}{\textbf{Claude}} & \multicolumn{3}{c}{\textbf{GPT}} & \multicolumn{3}{c}{\textbf{Gemini}} \\
\cmidrule(lr){2-4} \cmidrule(lr){5-7} \cmidrule(lr){8-10}
\textbf{Method} & EN & ES & Gap & EN & ES & Gap & EN & ES & Gap \\
\midrule
\multicolumn{10}{c}{1B} \\
Original (raw LL) & 33.3 & 30.9 & +2.5 & 33.3 & 30.9 & +2.5 & 33.3 & 30.9 & +2.5 \\
ParaEval (raw LL) & 34.0 & 33.5 & +0.5 & 34.9 & 34.7 & +0.2 & 35.3 & 34.9 & +0.4 \\
Paraphrasing correct (raw LL) & 62.5 & 65.4 & -2.9 & 62.2 & 63.7 & -1.4 & 70.5 & 70.0 & +0.5 \\
\midrule
\midrule
\multicolumn{10}{c}{3B} \\
Original (raw LL) & 37.3 & 33.5 & +3.8 & 37.3 & 33.5 & +3.8 & 37.3 & 33.5 & +3.8 \\
ParaEval (raw LL) & 38.3 & 37.8 & +0.5 & 38.9 & 38.5 & +0.4 & 38.8 & 38.4 & +0.4 \\
Paraphrasing correct (raw LL) & 66.1 & 67.3 & -1.2 & 65.3 & 66.5 & -1.2 & 73.0 & 72.0 & +1.0 \\
\midrule
\midrule
\multicolumn{10}{c}{8B} \\
Original (raw LL) & 41.2 & 37.8 & +3.4 & 41.2 & 37.8 & +3.4 & 41.2 & 37.8 & +3.4 \\
ParaEval (raw LL) & 42.1 & 41.7 & +0.4 & 42.7 & 42.4 & +0.2 & 42.3 & 41.9 & +0.4 \\
Paraphrasing correct (raw LL) & 69.6 & 70.3 & -0.7 & 68.4 & 69.3 & -0.9 & 75.4 & 73.6 & +1.9 \\
\bottomrule
\end{tabular}}
\end{table}

\begin{table}[t]
\centering
\caption{ARC-Easy diversified choices evaluation (1B, 5-shot). Orig = original translated EN, BT = back-translated EN, on the original english dataset (and paraphrased).}
\label{tab:bt-arce-diversified}
\resizebox{\textwidth}{!}{
\begin{tabular}{l ccc ccc ccc}
\toprule
 & \multicolumn{3}{c}{\textbf{Claude}} & \multicolumn{3}{c}{\textbf{GPT}} & \multicolumn{3}{c}{\textbf{Gemini}} \\
 \cmidrule(lr){2-4} \cmidrule(lr){5-7} \cmidrule(lr){8-10}
\textbf{Method} & Orig & BT & Gap & Orig & BT & Gap & Orig & BT & Gap \\
\midrule
Original (raw LL) & 72.3 & 70.0 & +2.4 & 72.3 & 70.0 & +2.4 & 72.3 & 70.0 & +2.4 \\
ParaEval (raw LL) & 70.4 & 70.3 & +0.1 & 70.4 & 70.5 & +0.1 & 71.1 & 69.9 & +1.2 \\
Paraphrasing correct (raw LL) & 79.9 & 79.1 & +0.8 & 81.0 & 80.9 & +0.1 & 79.7 & 78.9 & +0.8 \\
\bottomrule
\end{tabular}}
\end{table}

\begin{table}[t]
\centering
\caption{HellaSwag diversified choices evaluation (1B, 5-shot). Orig = original translated EN, BT = back-translated EN.}
\label{tab:bt-hswag-diversified}
\resizebox{\textwidth}{!}{
\begin{tabular}{l ccc ccc ccc}
\toprule
 & \multicolumn{3}{c}{\textbf{Claude}} & \multicolumn{3}{c}{\textbf{GPT}} & \multicolumn{3}{c}{\textbf{Gemini}} \\
 \cmidrule(lr){2-4} \cmidrule(lr){5-7} \cmidrule(lr){8-10}
\textbf{Method} & Orig & BT & Gap & Orig & BT & Gap & Orig & BT & Gap \\
\midrule
Original (raw LL) & 44.5 & 40.1 & +4.4 & 44.5 & 40.1 & +4.4 & 44.5 & 40.1 & +4.4 \\
ParaEval (raw LL) & 41.5 & 42.7 & -1.2 & 42.8 & 43.6 & -0.8 & 44.1 & 43.5 & +0.6 \\
Paraphrasing correct (raw LL) & 50.2 & 50.1 & +0.1 & 59.0 & 57.6 & +1.4 & 48.0 & 44.6 & +3.4 \\
\bottomrule
\end{tabular}}
\end{table}

\subsection{Frontier Model Evaluation}\label{app:frontier}
We evaluate four open-weight frontier models without retraining to test whether surface-form sensitivity persists at scale: LLaMA-3.1-70B, Qwen2.5-72B, Nemotron-3-120B (all pretrained-only), and GPT-OSS-120B (instruction-tuned). All models are evaluated 5-shot on English and Spanish ARC-Easy, ARC-Challenge, and HellaSwag using the same Claude-generated paraphrases as our controlled experiments.

\Cref{tab:oss-diversified_results} summarizes standard vs.\ ParaEval performance. Standard scoring finds EN--ES gaps of 3.0--7.3 on ARC-Challenge and 4.7--6.9 on HellaSwag. ParaEval reduces these by 14--32\% (0.9--1.4 absolute). ParaEval reduces gaps by bringing scores closer together, typically by lowering the higher-scoring language and raising the lower-scoring one, consistent with removing surface-form advantage. \Cref{tab:oss-diversified} shows the upper bound from paraphrasing only the correct answer, which inflates scores by 5--9, confirming that evaluating all options is necessary to avoid optimistic bias.

\subsection{Human Evaluation of Paraphrases}
We asked authors who are fluent or native in each language to annotate 20 paraphrases as keeping the same meaning and accurately replying to the question, or not.
We present the results in \Cref{tab:appendix/annotator_agreement}, in English there is a failure case, where the original question is "Which two terms are used to describe weather?", one of the original choices was "wind direction and amount of erosion", and the paraphrase is "how much erosion occurs and which way the wind blows", both annotators commented the question specifically asked for two terms, even though the paraphrase is correct, it does not keep \textit{exactly} the original meaning as per the question.
\begin{table}[]
    \caption{Agreement between annotators on 20 randomly select paraphrases for each language, comparing to the original and assessing whether it still answers the question correctly.}
    \label{tab:appendix/annotator_agreement}
    \centering
    \begin{tabular}{c|c|c|c|c}
        \toprule
        Language &  Annotator 1 &  Annotator 2 & Agreement & Correctness \\
        \midrule
        English & 19 & 19 & 100 & 95 \\
        Spanish & 20 & 20 & 100 & 100 \\
        \bottomrule
    \end{tabular}
\end{table}

\subsection{Effect of the number of paraphrases} \label{appendix:number_of_paraphrases}
\begin{figure}[ht]
    \centering
    \includegraphics[width=0.5\linewidth]{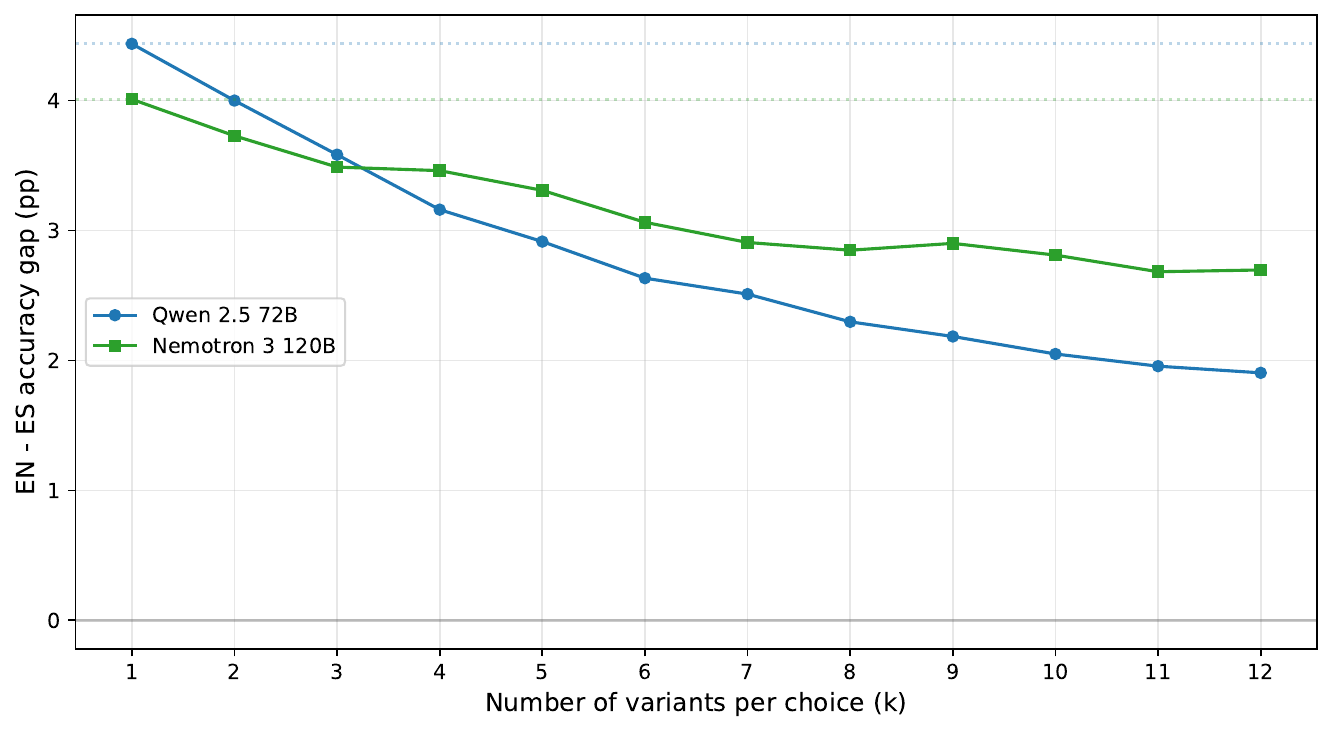}
    \caption{EN--ES accuracy gap on ARC-Challenge (raw LL) as a function of the number of paraphrase variants per choice ($k$) for Qwen\,2.5\,72B and Nemotron\,3\,120B. For each $k > 1$, we randomly sample $k{-}1$ paraphrases from a pool generated by three LLMs (Claude, GPT and Gemini), compute ParaEval accuracy, and average over 50 random draws. Dotted horizontal lines indicate the baseline gap ($k{=}1$). Both models exhibit a consistent monotonic decrease in the cross-lingual gap as $k$ increases, with Qwen\,2.5\,72B dropping from 4.4 to 1.9 (57\%) and Nemotron\,3\,120B from 4.0 to 2.7 (33\%), confirming that the trend observed at smaller scales (1B--8B) extends to large pretrained models.}
\label{fig:arcc-gap-vs-k-qwen-nemotron}
\end{figure}

In \Cref{fig:gap_vs_k_combined}a we show how the gap progresses with more alternatives for ARC-E, and in \Cref{fig:gap_vs_k_combined}b for ARC-C across all 3 scales.
We additionally check this affects models of all scales by additionally checking the impact in the larger Qwen and Nemotron models.%
\Cref{fig:arcc-gap-vs-k-qwen-nemotron} demonstrates that increasing $k$ monotonically reduces gaps for Qwen2.5-72B (4.4$\rightarrow$1.9) and Nemotron-3-120B (4.0$\rightarrow$2.7), mirroring the trend in \Cref{fig:gap_vs_k_combined} for smaller models. Together, these results establish that phrasing sensitivity is not a small-model artifact and that ParaEval generalizes to frontier-scale models without modification.

\begin{table}[t]
\centering
\caption{Diversified paraphrase evaluation on multilingual OSS models (5-shot). We report raw LL and ParaEval (best paraphrase per choice) accuracy for EN and ES, along with the EN--ES gap and its reduction under oracle scoring. All models but GPT-OSS are pretrained-only models.}
\label{tab:oss-diversified_results}
\resizebox{\textwidth}{!}{
\begin{tabular}{ll ccc ccc c}
\toprule
& & \multicolumn{3}{c}{\textbf{Original (raw LL)}} & \multicolumn{3}{c}{\textbf{ParaEval (raw LL)}} & \\
\cmidrule(lr){3-5} \cmidrule(lr){6-8}
\textbf{Model} & \textbf{Benchmark} & EN & ES & $\Delta_o$ & EN & ES & $\Delta_p$ & $\Delta_o - \Delta_p$ \\
\midrule
\multirow{3}{*}{GPT-OSS 120B} & ARC-Easy & 84.7 & 79.7 & +5.0 & 84.3 & 80.6 & +3.7 & +1.4 \\
 & ARC-C & 57.4 & 45.2 & +12.2 & 57.9 & 48.0 & +9.9 & +2.3 \\
  & MMLU & 48.6 & 41.4 & +7.2 & 50.0 & 43.3 & +6.7 & +0.5 \\
\midrule
\multirow{4}{*}{Qwen2.5 72B} & ARC-Easy & 89.4 & 86.4 & +3.0 & 89.1 & 87.0 & +2.2 & +0.8 \\
 & ARC-C & 65.8 & 61.3 & +4.4 & 66.2 & 63.1 & +3.1 & +1.4 \\
 & HellaSwag & 65.4 & 58.5 & +6.9 & 62.6 & 56.6 & +6.0 & +0.9 \\
  & MMLU & 54.6 & 47.3 & +7.3 & 55.4 & 51.3 & +4.1 & +3.2 \\
\midrule
\multirow{4}{*}{LLaMA-3.1 70B} & ARC-Easy & 89.7 & 85.8 & +3.9 & 88.9 & 86.2 & +2.7 & +1.2 \\
 & ARC-C & 66.2 & 58.9 & +7.3 & 66.5 & 60.2 & +6.3 & +1.0 \\
 & HellaSwag & 64.7 & 58.6 & +6.1 & 61.6 & 56.6 & +5.0 & +1.1 \\
 & MMLU & 53.1 & 48.1 & +5.0 & 54.1 & 49.6 & +4.5 & +0.4 \\
\midrule
\multirow{4}{*}{Nemotron-3 120B} & ARC-Easy & 88.2 & 86.5 & +1.7 & 88.3 & 88.0 & +0.3 & +1.4 \\
 & ARC-C & 66.5 & 62.5 & +4.0 & 67.1 & 64.3 & +2.7 & +1.3 \\
 & HellaSwag & 68.0 & 63.3 & +4.7 & 64.7 & 61.4 & +3.4 & +1.4 \\
& MMLU & 58.4 & 53.3 & +5.0 & 59.7 & 56.5 & +3.2 & +1.9 \\
\bottomrule
\end{tabular}}
\end{table}

\begin{table}[t]
\centering
\caption{Diversified paraphrase evaluation on multilingual OSS models (5-shot, Claude paraphrases). We compare original raw LL scoring with oracle on the correct answer only.}
\label{tab:oss-diversified}
\resizebox{\textwidth}{!}{
\begin{tabular}{ll cc cc}
\toprule
& & \multicolumn{2}{c}{\textbf{Original (raw LL)}} & \multicolumn{2}{c}{\textbf{Paraphrasing Correct (raw LL)}} \\
\cmidrule(lr){3-4} \cmidrule(lr){5-6}
\textbf{Model} & \textbf{Benchmark} & EN & ES & EN & ES \\
\midrule
\multirow{3}{*}{GPT-OSS 120B} & ARC-Easy & 84.7 & 79.7 & 90.4 & 87.2 \\
 & ARC-C & 57.4 & 45.2 & 66.4 & 57.4 \\
 & MMLU & 48.6 & 41.4 & 61.4 & 63.0 \\
\midrule
\multirow{4}{*}{Qwen2.5-72B} & ARC-Easy & 89.4 & 86.4 & 93.1 & 91.2 \\
 & ARC-C & 65.8 & 61.3 & 71.8 & 70.7 \\
 & HellaSwag & 65.4 & 58.5 & 68.8 & 62.8 \\
 & MMLU & 54.6 & 47.3 & 66.1 & 66.1 \\
\midrule
\multirow{4}{*}{LLaMA-3.1-70B} & ARC-Easy & 89.7 & 85.8 & 93.1 & 91.2 \\
 & ARC-C & 66.2 & 58.9 & 72.4 & 68.3 \\
 & HellaSwag & 64.7 & 58.6 & 69.2 & 62.7 \\
 & MMLU & 53.1 & 48.1 & 66.0 & 67.9 \\
\midrule
\multirow{4}{*}{Nemotron-3 120B} & ARC-Easy & 88.2 & 86.5 & 91.9 & 90.7 \\
 & ARC-C & 66.5 & 62.5 & 73.5 & 71.2 \\
 & HellaSwag & 68.0 & 63.3 & 71.4 & 67.0 \\
  & MMLU & 58.4 & 53.3 & 66.9 & 69.2 \\
\bottomrule
\end{tabular}}
\end{table}